\documentclass[5pt]{article}

\usepackage[letterpaper]{geometry}
\usepackage{spconf,amsmath,epsfig}
\usepackage{enumitem}

\usepackage{fancyhdr}
\renewcommand{\headrulewidth}{0pt}
\usepackage{amsmath,spconf,epsfig}
\usepackage{amssymb,amsfonts,amsthm}
\usepackage{caption}
\usepackage{booktabs}
\usepackage{tabularx}
\usepackage[table]{xcolor}
\usepackage{cite}
\usepackage{epstopdf}
\usepackage{graphics,caption,subcaption}
\usepackage{xcolor}
\usepackage[export]{adjustbox}
\definecolor{StruckCol}{RGB}{232.0500  104.5500   43.3500}
\definecolor{MILCol}{RGB}{ 160 32 240}
\definecolor{Brownish}{RGB}{ 139 69 19}
\definecolor{vilt}{RGB}{255 28 174}

\usepackage{tikz,graphicx}
\usepackage{array}
\usepackage{tabu}
\usepackage{CJKutf8}
\usepackage{setspace,lipsum}
\DeclareMathOperator*{\argmin}{\arg\!\min}

\newcommand{\Cbf}{\mathbf{C}}
\newcommand{\cbf}{\mathbf{c}}

\newcommand{\Dbf}{\mathbf{D}}

\newcommand{\Ebf}{\mathbf{E}}

\newcommand{\Gbf}{\mathbf{G}}

\newcommand{\Kcal}{\mathcal{K}}

\newcommand{\Ncal}{\mathcal{N}}

\newcommand{\Rbb}{\mathbb{R}}

\newcommand{\xbf}{\mathbf{x}}
\newcommand{\Xbf}{\mathbf{X}}

\newcommand{\Ybf}{\mathbf{Y}}
\newcommand{\Ycal}{\mathcal{Y}}

\newcommand{\Zbf}{\mathbf{Z}}

\begin{document}\sloppy
\topmargin=0mm
\def\x{{\mathbf x}}
\def\L{{\cal L}}

\title{Robust Structured Multi-task Multi-view Sparse Tracking}
%
\name{Mohammadreza Javanmardi, Xiaojun Qi }
\address{Department of Computer Science, Utah State University, Logan,
UT, 84322 USA \\
javanmardi@aggiemail.usu.edu, xiaojun.qi@usu.edu}
%
%
%

\maketitle

\begin{abstract}
Sparse representation is a viable solution to visual tracking. In this paper, we propose a structured multi-task multi-view tracking (SMTMVT) method, which exploits the sparse appearance model in the particle filter framework to track targets under different challenges. Specifically, we extract features of the target candidates from different views and sparsely represent them by a linear combination of templates of different views. Unlike the conventional sparse trackers, SMTMVT not only jointly considers the relationship between different tasks and different views but also retains the structures among different views in a robust multi-task multi-view formulation. We introduce a numerical algorithm based on the proximal gradient method to quickly and effectively find the sparsity by dividing the optimization problem into two subproblems with the closed-form solutions. Both qualitative and quantitative evaluations on the benchmark of challenging image sequences demonstrate the superior performance of the proposed tracker against various state-of-the-art trackers.
\end{abstract}
\section{Introduction}
\label{sec:intro}\vspace{-3pt}
Visual tracking is the process of estimating states of a moving target in a dynamic frame sequence. It is considered as one of the most important and challenging topics in computer vision and has overabundant applications in surveillance, human motion analysis, smart vehicles transportation, navigation, etc. Although numerous methods \cite{yilmaz2006object,salti2012adaptive,kristan2015visual} have been introduced in recent years, it is still challenging to develop a robust tracking algorithm due to occlusion, illumination variations, deformation, camera motion, background clutter, etc. 

Tracking algorithms can be classified into two categories: discriminative and generative. Discriminative approaches formulate a decision boundary to separate the target from the background. For example, Avidan \cite{avidan2007ensemble} proposes the ensemble tracking method, which combines a set of weak classifiers into a strong one to label the candidate as target or background. Grabner et al. \cite{grabner2008semi} propose semi-supervised boosting to alleviate the drifting problem in tracking. Bebenko et al. \cite{babenko2009visual} use a large number of positive and negative bags consisting of image patches to update a multiple instance learning-based appearance model. In contrast, generative approaches adopt a model to represent the target and formulate the tracking as a model-based searching procedure to find the most similar region to the target. Black et al. \cite{black1998eigentracking} formulate the eigenspace model to represent the target and employ a coarse-to-fine matching strategy to track the target over image sequences. Adam et al. \cite{adam2006robust} propose the Frag-Track algorithm to track an object, which represents a template object using multiple arbitrary image patches and creates the vote map using the integral histogram. Ross et al. \cite{ross2008incremental} learn a low-dimensional subspace representation of the target to track objects.

Sparse representation based trackers (sparse trackers) are considered as the generative tracking methods since they express features of a target as a sparse linear combination of a template set. Based on the number of employed features, sparse trackers are further classified into single-view and multi-view. Single-view sparse tracking approaches represent one feature (i.e., pixel intensity) of a target region  using a set of templates. Mei et al. \cite{mei2011robust} propose the L1T tracker, which represents the intensity of each target candidate by a set of templates and finds the sparsity by solving an $\ell_1$ minimization problem. Li et al. \cite{li2011real} propose to adopt the orthogonal matching pursuit (OMP) algorithm \cite{pati1993orthogonal} to reduce the complexity of solving the minimization problem in L1T tracker. The modified version of OMP \cite{shekaramiz2015block} may be incorporated to further reduce the complexity. Zhang et al. \cite{zhang2012robust} propose to jointly learn the intensities of all target candidates. These methods adopt a global sparse appearance model to represent a target as an entity region. Therefore, they are less effective in handling large occlusion. To address this problem, Jia et al. \cite{jia2012visual} divide each candidate into a set of overlapping patches and represent them by a template set of the patches. Zhang et al. \cite{zhang2015structural} exploit the global and local representation of a target candidate to achieve robust tracking. These methods solve the problems associated with occlusion and noise. However, they are sensitive to shape deformation of targets and varied illumination due to the use of intensity values. In contrast, multi-view sparse tracking approaches extract visual features such as color, edge, texture, and histogram to complement the intensity of the target. Exploiting multi-view information has also been widely used in many computer vision tasks such as visual classification, face recognition, and image segmentation \cite{yuan2012visual,zohrizadeh2016reliability}. For instance, Zohrizadeh et al. \cite{zohrizadeh2016reliability} employ multiple local features in a non-negative matrix factorization framework to segment an image. Hong et al. \cite{hong2013tracking} propose a tracker that considers the underlying relationship among different views and particles in terms of the least square (LS). To handle the data contaminated by outliers and noise, Mei et al. \cite{mei2015robust} use the least absolute deviation (LAD) in their optimization model. Both approaches cannot retain the underlying layout structure among different views. In other words, different views of a target candidate may be reconstructed by activating the same templates in the dictionary set, whose representation coefficients do not resemble the similar combination of activated templates.

To address these issues, we propose a novel structured multi-task multi-view tracking (SMTMVT) method to track objects under different challenges. Similar to \cite{hong2013tracking}, SMTMVT exploits multi-view information such as intensity, edge, and histogram of target candidates and jointly represents them using templates. However, SMTMVT improves Hong's tracker by proposing a new optimization model to attain the underlying layout structure among different views and reduce the error corresponding to outlier target candidates. The main contributions of the proposed work are summarized as: 1) Designing a novel optimization model to effectively utilize a nuclear norm of the sparsity for multi-task multi-view sparse trackers; 2) Representing a particular view of a target candidate as an individual task and simultaneously retaining the underlying layout structure among different views; 3) Incorporating an outlier minimization term in the optimization model to efficiently reduce the error of outlier target candidates; 4) Adopting the proximal gradient (PG) method to quickly and effectively solve the optimization problem. 

The remainder of this paper is as follows: Section \ref{sec:not} introduces the notations.  Section \ref{sec:pro} presents the SMTMVT method together with its optimization model solved by our proposed PG-based numerical algorithm.. Section \ref{sec:exp} demonstrates the experimental results on 15 publicly challenging image sequences and the CVPR2013 tracking benchmark and compares the results of the proposed method with several state-of-the-art methods. Section \ref{sec:conc} draws the conclusions.
\noindent\section{Notations}\vspace{-10pt}\label{sec:not}
Throughout this paper, we use bold lowercase and bold uppercase letters to denote vectors and matrices, respectively. Specifically, two sets of numbers $\{1,2,\dots,n\}$ and $\{1,2,\dots,K\}$ are respectively denoted by $\Ncal$ and $\Kcal$. Vector $\mathbf{1}$ is a column vector of all ones of an appropriate dimension. For a given matrix $\Ybf\in\Rbb^{n\times m}$, we denote its Frobenius norm, nuclear norm, and $L_1$ norm by ${\left\|\Ybf\right\|}_F$, ${\left\|\Ybf\right\|}_*$, and ${\left\|\Ybf\right\|}_1$, respectively. The \textit{soft-thresholding operator} is defined as $S_{\rho}(\Ybf) =\sum_{i=1}^{n}\sum_{j=1}^{m} \text{sign}(\Ybf_{ij})\,\text{max}(|\Ybf_{ij}|-\rho, 0)$. For a set $\Ycal$, the indicator function $\delta_{\mathcal{Y}}(\Ybf)$ returns $+\infty$ when $\Ybf\notin\Ycal$ and returns $0$ when $\Ybf\in\Ycal$. The \textit{proximal operator} is defined as $\text{Prox}_{\sigma}^{f}(\Ybf)= \argmin_{\Zbf} f(\Zbf)+\frac{1}{2\sigma}{\left\| \Zbf-\Ybf\right\|}_{F}^{2}$, where $\sigma>0$ and $f(\cdot)$ is a given function.

\section{The Proposed SMTMVT Method}\label{sec:pro}
This section provides detailed information about the proposed particle filter based tracker. Specifically, we formulate a  sparse appearance model in the proposed SMTMVT and propose a numerical solution to efficiently solve the model. 
\vspace{-3pt}
\subsection{Structured Multi-Task Multi-View Tracking (SMTMVT)}\vspace{-3pt}

The proposed SMTMVT method utilizes the sparse appearance model to exploit multi-task multi-view information in a new optimization model, attain the underlying layout structure among different views, and reduce the error of outlier target candidates. At time $t$, we consider $n$ particles with their corresponding image observations (target candidates). Using the state of the $i$-th particle, its observation is obtained by cropping the region of interest around the target. Each observation is considered to have $K$ different views. For the $k$-th view, $d_k$ dimensional feature vectors of all particles, $\{\xbf_i^k\}_{i\in\Ncal}$, are combined to form the matrix $\Xbf^k = [\xbf_1^k,\dots,\xbf_n^k] \in \Rbb^{d_k\times n}$ and $N$ target templates are used to create its target dictionary $\Dbf^k \in \Rbb^{d_k\times N}$.
Following the same notations in \cite{hong2013tracking}, we use the $k$-th dictionary $\Dbf^k$ to represent the $k$-th feature matrix $\Xbf^k$ and learn the sparsity $\Cbf^k \in \Rbb^{N\times n}$. In addition, We divide the reconstruction errors of the $k$-th view into two components as follows:
\begin{equation}
\Xbf^{k}-\Dbf^{k} \Cbf^{k} = \Gbf^{k} + \Ebf^{k}
\end{equation}
The first error component $\Gbf^{k} \in \Rbb^{d_k\times n}$ corresponds to the minor reconstruction errors resulted from the representation of good target candidates. The second error component $\Ebf^{k} \in \Rbb^{d_{k}\times n}$ corresponds to the significant reconstruction errors resulted from the representation of outlier target candidates. We use the Frobenius norm factor minimization of $\Gbf^{k}$ error to minimize the square root of the sum of the absolute squares of its elements and adopt the $\ell_1$ norm minimization of $\Ebf^{k}$ error to minimize the maximum column-sum of its elements. This assures the reconstruction errors for both good and bad target candidates are minimized.

To boost the performance, we maintain the underlying layout structure between $K$ different views. For the $i$-th particle, we not only represent all its feature vectors $\{\xbf_i^k\}_{k\in\Kcal}$ by activating the same subset of target templates in the target dictionaries (i.e., $\{\Dbf^k\}_{k\in\Kcal}$), but also equalize the representation coefficients of activated templates for all $K$ views. In other words, we aim to resemble the $i$th columns of $\Cbf^{k}$s, for $k\in\Kcal$, to have a similar representation structure in terms of the activated templates and similar coefficients in terms of the activated values. To do so, we concatenate $\Cbf^{k}$s to form $\Cbf \in \Rbb^ {N \times (nK)}$, which is the sparsity matrix corresponding to the representation of $K$ views in $n$ observations. We then minimize the nuclear norm of the matrix $\Pi_{i,n}(\Cbf)$, which is a good surrogate of the rank minimization, to ensure the columns to be similar or linearly dependent of each other. Here, $\Pi_{i,n}(\Cbf)$ selects a sub-matrix of columns of $\Cbf$, whose index belongs to the set $\{(l-1)n+i\}_{l\in\Kcal}$. The selected columns are the simultaneous columns of the $i$-th target candidate in different views.

We formulate the SMTMVT sparse appearance model as the following optimization problem by jointly evaluating its $K$ view matrices $\{\Xbf_i^k\}_{k\in\Kcal}$ with $n$ different particles (tasks):
\begin{subequations}\label{eq:generalOPT}
\begin{align}
&\hspace{-0.25cm} \underset{\begin{subarray}{l}\Cbf,\Ebf,\{\Gbf_k\}\end{subarray}}{\text{minimize}}
&&\hspace{-0.18cm} {\sum_{k\in\Kcal}\!{\left\|\Gbf^k\right\|}_F^2}\!+\!{\lambda\!\sum_{i\in\Ncal}\! {\left\|\Pi_{i,n}(\Cbf)\right\|}_*}\!+\!\gamma ({\left\|\Cbf\right\|}_1\!+\!{\left\|\Ebf\right\|}_1)\label{eq:generalOPT_con0}\\ 
&\hspace{-0.25cm} {\text{subject to}}
&&\hspace{-0.18cm}\Xbf^k=\Dbf^k\Cbf^k+\Gbf^k + \Ebf^k,~~~~k\in\Kcal\label{eq:generalOPT_con1}\\
&&&\hspace{-0.18cm}\Cbf\geq 0 \label{eq:generalOPT_con2}
\end{align}
\end{subequations}
where $\textbf{E}^{k}$'s are vertically stacked to form the bigger matrix $\Ebf \in \Rbb ^ {(\sum_{k\in\Kcal}d_k)\times n}$, parameter $\lambda$ regularizes the nuclear norm of $\Cbf$, and parameter $\gamma$ controls the sparsity of $\Cbf$ and $\Ebf$.

Due to the equality constraint \eqref{eq:generalOPT_con1}, we eliminate $\Gbf^k$ from \eqref{eq:generalOPT_con0} and equivalently solve the following problem:
\begin{subequations}\label{eq:generalOPT2}
\begin{align}
&\!\!\underset{\begin{subarray}{l}\Cbf,\Ebf\end{subarray}}{\text{minimize}}
&&\!\! {\sum_{k\in\Kcal} {\left\|\Xbf^k-\Dbf^k\Cbf^k-\Ebf^k\right\|}_F^2}+\!{\lambda \sum_{i\in\Ncal} {\left\|\Pi_{i,n}(\Cbf)\right\|}_*}\nonumber\\
&&&\!\!\! +\!\gamma (\mathbf{1}^{\!\top}\!\Cbf\mathbf{1}+{\left\|\Ebf\right\|}_1)\\
&{\!\!\text{subject to}}
&&\Cbf\geq 0 \label{eq:generalOPT2_con1}
\end{align}
\end{subequations}

Finally, we compute the likelihood of the $i$-th candidate as follows: 
\begin{equation} 
p_i=\exp(-\alpha\sum_{k\in\Kcal} {\left\|\Dbf^k\cbf_i^k-\xbf_i^k\right\|}^2)
\end{equation}
where $\cbf_i^k$ is the sparse coefficients of the $i$-th candidate corresponding to the target templates in the $k$-th view and $\alpha$ is a constant value. We select the candidate with the highest likelihood value as the tracking result at frame $t$. Similar to \cite{mei2011robust}, we update the target templates to handle the appearance changes of the target throughout the frame sequences.
\vspace{-13pt}
\subsection{Optimization Algorithm}\vspace{-5pt}
Since the convex problem in \eqref{eq:generalOPT2} can be split into differentiable and non-differentiable subproblems, we adopt the PG method \cite{parikh2014proximal} to develop a numerical solution to the proposed model. To do so, we cast the differentiable subproblem as follows:
\begin{equation}\label{eq:lagrang}
\begin{aligned}
\mathcal{L}(\Cbf,\Ebf) = &{\sum_{k\in\Kcal} {\left\|\Xbf^k-\Dbf^k\Cbf^k-\Ebf^k\right\|}_F^2}+{\lambda \sum_{i\in\Ncal} {\left\|\Pi_{i,n}(\Cbf)\right\|}_*}\\
&\!+\!\gamma\,(\mathbf{1}^{\!\top}\!\Cbf\mathbf{1})
\end{aligned}
\end{equation}
This equation is sub-differentiable with respect to $\Cbf$ and differentiable with respect to $\Ebf$. Hence, two variables $\Cbf$ and $\Ebf$ can be updated at time $t+1$ by the following equations:
\begin{subequations}\label{eq:prox}
\begin{align}
&\Ebf^{(t+1)} := \text{Prox}_{\sigma}^{f_{\Ebf}}(\Ebf^{(t)}-\sigma\,\nabla\!\mathcal{L}(\Cbf^{(t)},\Ebf^{(t)})),\label{eq:prox_E}\\
&\Cbf^{(t+1)} := \text{Prox}_{\sigma}^{f_{\Cbf}}(\Cbf^{(t)}-\sigma\,\nabla\!\mathcal{L}(\Cbf^{(t)},\Ebf^{(t)})),\label{eq:prox_C}
\end{align}
\end{subequations}
where the step-size $\sigma$ controls the convergence rate, $\nabla\!\mathcal{L}(\cdot,\cdot)$ is the sub-gradient operator, $f_{\Ebf} = {\left\|\Ebf\right\|}_1$, and $f_{\Cbf}=\delta_{\mathcal{C}}$. We adopt the computationally efficient PG algorithm to iteratively update $\Ebf$ and $\Cbf$, which are initially set as 0's, until they converge to the constant matrices. Both subproblems \eqref{eq:prox_E} and \eqref{eq:prox_C} can be easily solved via the existing methods. Specifically, \eqref{eq:prox_E} is a $\ell_{1}$-minimization problem with an analytic solution, which can be obtained using soft thresholding, i.e., $S_{(\sigma \gamma)}(\cdot)$. Moreover, \eqref{eq:prox_C} is an Euclidean norm projection onto the nonnegative orthant, which enjoys the closed-form solution.  It should be emphasized that the convergence rate of this numerical algorithm can be further improved by the acceleration techniques presented in \cite{nesterov2005smooth}.
\section{Experimental Results}\vspace{-3pt}\label{sec:exp}
In this section, we evaluate the performance of the proposed method on 15 publicly available frame sequences and the CVPR2013 tracking benchmark data set \cite{wu2013online}. 

To ensure fair comparison, We employ four popular features as used in \cite{hong2013tracking,mei2015robust} in the proposed SMTMVT method. These features are intensity, color histogram, histogram of oriented gradients (HOGs) \cite{dalal2005histograms}, and local binary patterns (LBPs) \cite{ojala2002multiresolution}.  In addition, we employ a simple but effective illumination normalization method \cite{tan2010enhanced} before feature extraction to eliminate the effect of illumination and improve the quality and discriminative power of the features. Following the same settings in \cite{hong2013tracking,mei2015robust}, we set the size of intensity template to be one third of the size of the initial target or the half size of the initial target when its shorter length is less than 20 pixels. For all the experiments, we set $\lambda\!=\!0.1$, $\gamma\!=\!0.25$, $\alpha\!=\!30$, the number of particles $n\!=\!400$, and the number of target templates $N\!=\!10$.
\vspace{-10pt}
\subsection{Experiments on Publicly Available Sequences }\vspace{-3pt}
We extensively conduct experiments on 15 challenging frame sequences and follow the same settings as in \cite{hong2013tracking,mei2015robust} to resize all frames to 320$\times$240. We compare the proposed SMTMVT method with eight state-of-the-art tracking methods, namely, L1 tracker \cite{mei2011robust}, multi-task tracking (MTT) \cite{zhang2012robust}, Struck \cite{Struck2011}, tracking with multiple instance learning (MIL) \cite{babenko2011robust}, incremental learning for visual tracking (IVT) \cite{ross2008incremental}, visual tracking decomposition (VTD) \cite{kwon2010visual}, multi-task multi-view tracking least square (MTMVTLS) \cite{hong2013tracking}, and multi-task multi-view tracking least absolute deviation (MTMVTLAD) \cite{mei2015robust}. We use the publicly available source code or binary code provided by the authors to produce the tracking results.  We use the default parameters for initialization.

\begin{figure*}[t]
\centering
\captionsetup{justification=centering}
\begin{adjustbox}{minipage=\textwidth,scale=.94}
\begin{subfigure}[normal]{0.328\textwidth}
\centering
		\begin{subfigure}[normal]{0.492\linewidth}
        \begin{tikzpicture}
        \node{\includegraphics[width=\linewidth,height=0.75\linewidth]{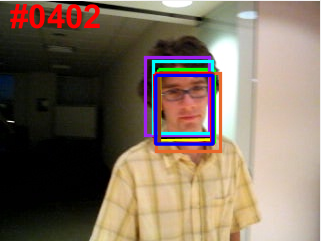}};
        \end{tikzpicture}
		\end{subfigure}
		\begin{subfigure}[normal]{0.492\linewidth}
			\begin{tikzpicture}
              \node {\includegraphics[width=\linewidth,height=0.75\linewidth]{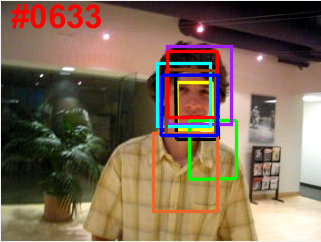}};
              \node [draw, fill=white, font=\bfseries, text=black] at (0.295\linewidth,0.28\linewidth) {\scriptsize david1};
             \end{tikzpicture}
		\end{subfigure}
\end{subfigure}
\begin{subfigure}[normal]{0.328\textwidth}
 \centering
 \begin{subfigure}[normal]{0.492\linewidth}
 	\begin{tikzpicture}
			\node{\includegraphics[width=\linewidth,height=0.75\linewidth]{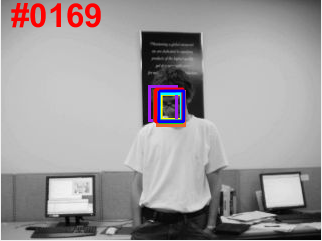}};
    \end{tikzpicture}
		\end{subfigure}
		\begin{subfigure}[normal]{0.492\linewidth}
			\begin{tikzpicture}
              \node {\includegraphics[width=\linewidth,height=0.75\linewidth]{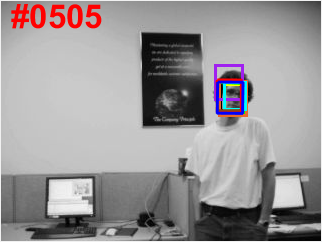}};
              \node [draw, fill=white, font=\bfseries, text=black] at (0.295\linewidth,0.28\linewidth) {\scriptsize david2};
             \end{tikzpicture}
		\end{subfigure}
\end{subfigure}
\begin{subfigure}[normal]{0.328\textwidth}
\centering
\begin{subfigure}[normal]{0.492\linewidth}
\begin{tikzpicture}
			\node{\includegraphics[width=\linewidth,height=0.75\linewidth]{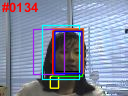}};
            \end{tikzpicture}
		\end{subfigure}
		\begin{subfigure}[normal]{0.492\linewidth}
			\begin{tikzpicture}
              \node {\includegraphics[width=\linewidth,height=0.75\linewidth]{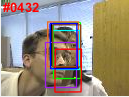}};
              \node [draw, fill=white, font=\bfseries, text=black] at (0.365\linewidth,0.28\linewidth) {\scriptsize girl};
             \end{tikzpicture}
		\end{subfigure}       
\end{subfigure}\vspace{-0.18cm}
\\ 
\begin{subfigure}[normal]{0.328\textwidth}
\centering
\begin{subfigure}[normal]{0.492\linewidth}
	\begin{tikzpicture}
		\node{\includegraphics[width=\linewidth,height=0.75\linewidth]{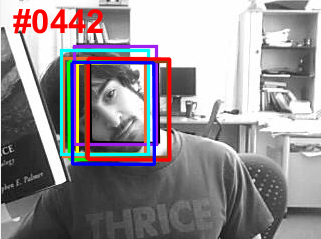}};
	\end{tikzpicture}
\end{subfigure}
\begin{subfigure}[normal]{0.492\linewidth}
	\begin{tikzpicture}
              \node {\includegraphics[width=\linewidth,height=0.75\linewidth]{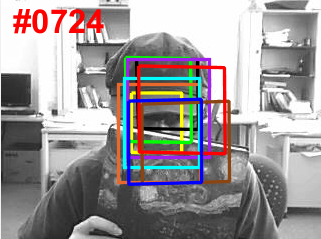}};
              \node [draw, fill=white, font=\bfseries, text=black] at (0.275\linewidth,0.285\linewidth) {\scriptsize faceocc2};
	\end{tikzpicture}
\end{subfigure}
\end{subfigure}
\begin{subfigure}[normal]{0.328\textwidth}
 \centering
 \begin{subfigure}[normal]{0.492\linewidth}
 \begin{tikzpicture}
			\node{\includegraphics[width=\linewidth,height=0.75\linewidth]{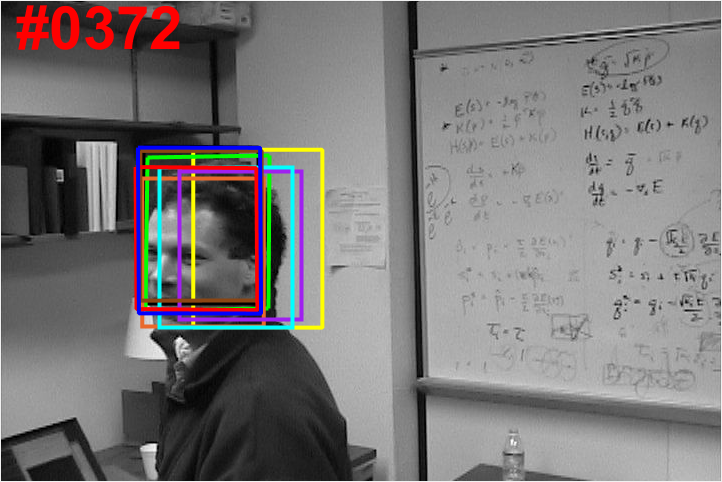}};
            \end{tikzpicture}
		\end{subfigure}
		\begin{subfigure}[normal]{0.492\linewidth}
			\begin{tikzpicture}
              \node {\includegraphics[width=\linewidth,height=0.75\linewidth]{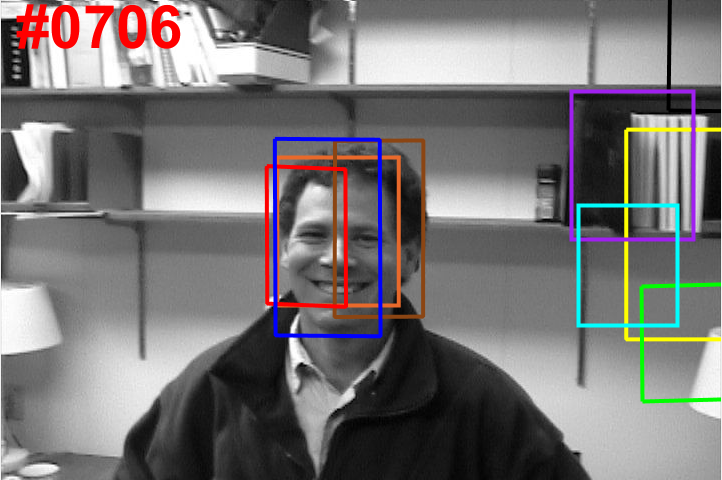}};
              \node [draw, fill=white, font=\bfseries, text=black] at (0.275\linewidth,0.285\linewidth) {\scriptsize fleetface};
             \end{tikzpicture}
		\end{subfigure}
\end{subfigure}
\begin{subfigure}[normal]{0.328\textwidth}
 \centering
 \begin{subfigure}[normal]{0.492\linewidth}
 \begin{tikzpicture}
			\node{\includegraphics[width=\linewidth,height=0.75\linewidth]{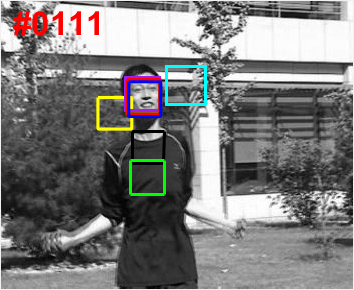}};
            \end{tikzpicture}
		\end{subfigure}
		\begin{subfigure}[normal]{0.492\linewidth}
			\begin{tikzpicture}
              \node {\includegraphics[width=\linewidth,height=0.75\linewidth]{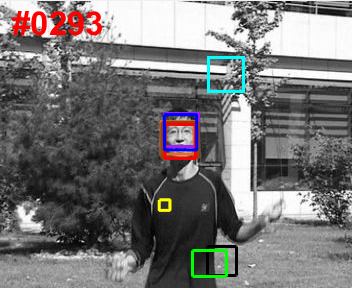}};
              \node [draw, fill=white, font=\bfseries, text=black] at (0.295\linewidth,0.285\linewidth) {\scriptsize jumping};
             \end{tikzpicture}
		\end{subfigure}
\end{subfigure}\vspace{-0.18cm}
\\
\begin{subfigure}[normal]{0.328\textwidth}
 \centering
 \begin{subfigure}[normal]{0.492\linewidth}
 	\begin{tikzpicture}
			\node{\includegraphics[width=\linewidth,height=0.75\linewidth]{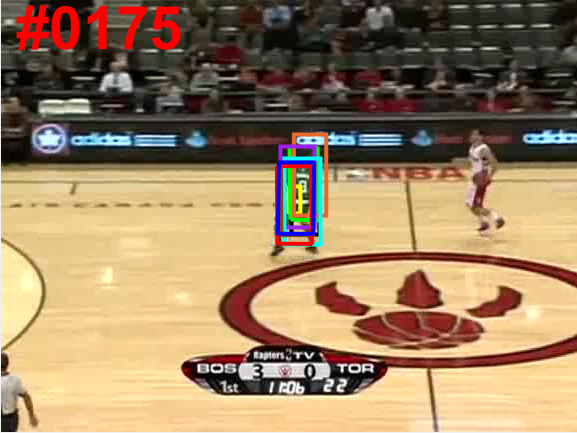}};
    \end{tikzpicture}
		\end{subfigure}
		\begin{subfigure}[normal]{0.492\linewidth}
			\begin{tikzpicture}
              \node {\includegraphics[width=\linewidth,height=0.75\linewidth]{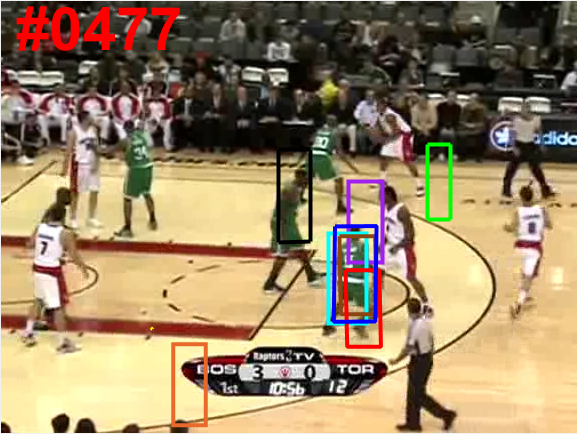}};
              \node [draw, fill=white, font=\bfseries, text=black] at (0.245\linewidth,0.295\linewidth) {\scriptsize basketball};
             \end{tikzpicture}
		\end{subfigure}
\end{subfigure}
\begin{subfigure}[normal]{0.328\textwidth}
 \centering
 \begin{subfigure}[normal]{0.492\linewidth}
 \begin{tikzpicture}
			\node{\includegraphics[width=\linewidth,height=0.75\linewidth]{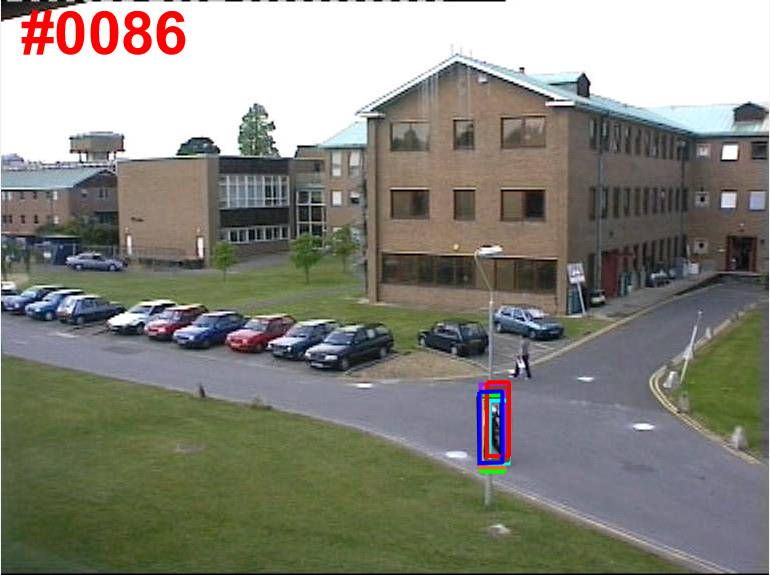}};
\end{tikzpicture}
		\end{subfigure}
		\begin{subfigure}[normal]{0.492\linewidth}
			\begin{tikzpicture}
              \node {\includegraphics[width=\linewidth,height=0.75\linewidth]{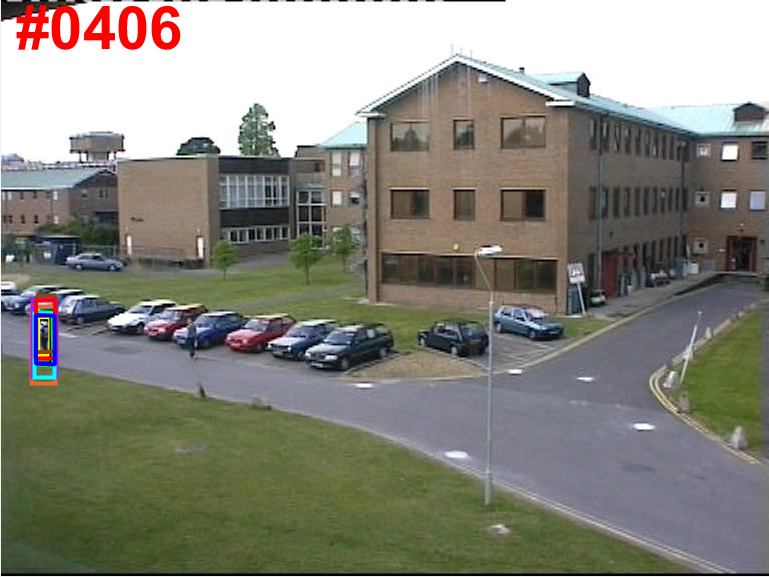}};
              \node [draw, fill=white, font=\bfseries, text=black] at (0.275\linewidth,0.285\linewidth) {\scriptsize walking};
             \end{tikzpicture}
		\end{subfigure}
\end{subfigure}
\begin{subfigure}[normal]{0.328\textwidth}
 \centering
 \begin{subfigure}[normal]{0.492\linewidth}
 \begin{tikzpicture}
			\node{\includegraphics[width=\linewidth,height=0.75\linewidth]{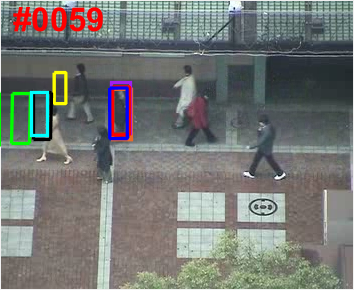}};
\end{tikzpicture}
		\end{subfigure}
		\begin{subfigure}[normal]{0.492\linewidth}
			\begin{tikzpicture}
              \node {\includegraphics[width=\linewidth,height=0.75\linewidth]{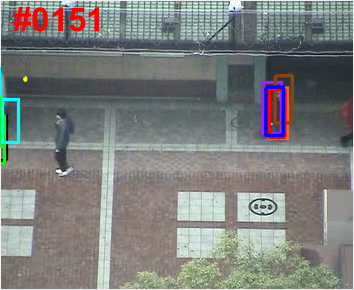}};
              \node [draw, fill=white, font=\bfseries, text=black] at (0.305\linewidth,0.290\linewidth) {\scriptsize subway};
             \end{tikzpicture}
		\end{subfigure}
\end{subfigure}\vspace{-0.18cm}
\\
\begin{subfigure}[normal]{0.328\textwidth}
 \centering
 \begin{subfigure}[normal]{0.492\linewidth}
			 \begin{tikzpicture}
            \node{\includegraphics[width=\linewidth,height=0.75\linewidth]{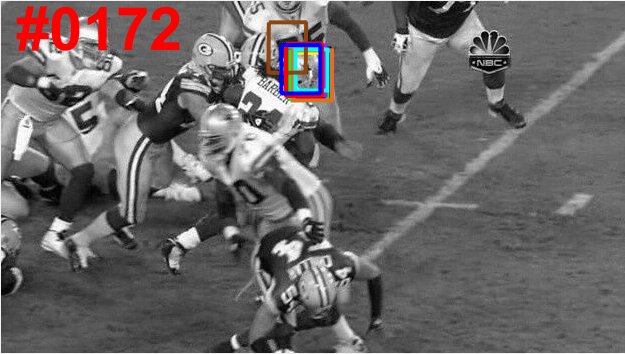}};
		    \end{tikzpicture}
        \end{subfigure}
		\begin{subfigure}[normal]{0.492\linewidth}
			\begin{tikzpicture}
              \node {\includegraphics[width=\linewidth,height=0.75\linewidth]{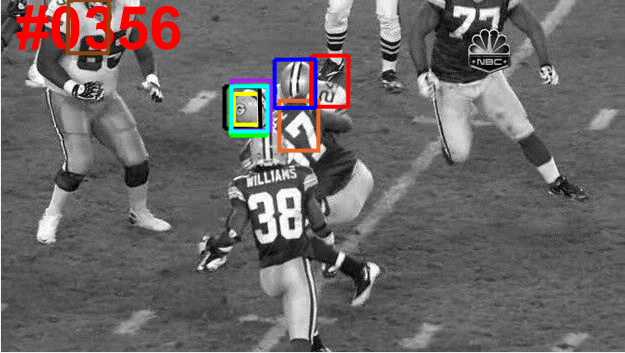}};
              \node [draw, fill=white, font=\bfseries, text=black] at (0.285\linewidth,0.295\linewidth) {\scriptsize football};
             \end{tikzpicture}
		\end{subfigure}
\end{subfigure}
\begin{subfigure}[normal]{0.328\textwidth}
 \centering
 \begin{subfigure}[normal]{0.492\linewidth}
			\begin{tikzpicture}
            \node{\includegraphics[width=\linewidth,height=0.75\linewidth]{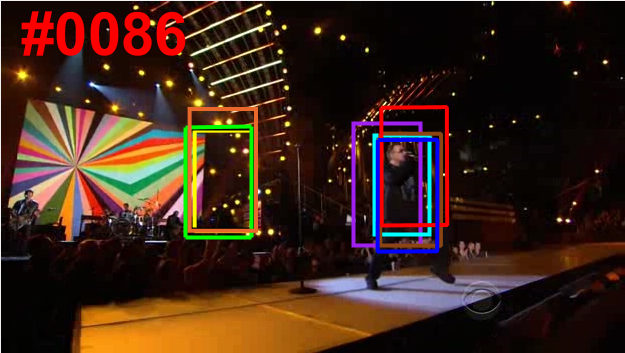}};
            \end{tikzpicture}
		\end{subfigure}
		\begin{subfigure}[normal]{0.492\linewidth}
			\begin{tikzpicture}
              \node {\includegraphics[width=\linewidth,height=0.75\linewidth]{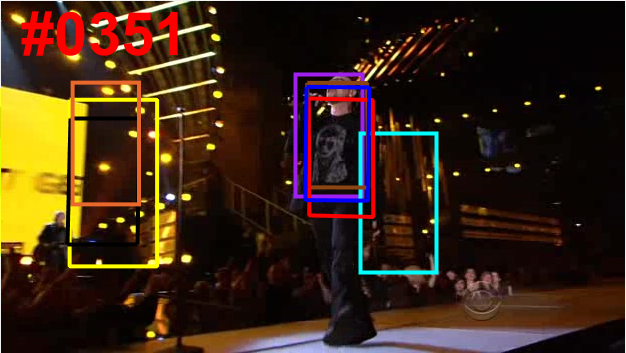}};
              \node [draw, fill=white, font=\bfseries, text=black] at (0.285\linewidth,0.295\linewidth) {\scriptsize singer2};
             \end{tikzpicture}
		\end{subfigure}
\end{subfigure}
\begin{subfigure}[normal]{0.328\textwidth}
 \centering
 \begin{subfigure}[normal]{0.492\linewidth}
			\begin{tikzpicture}
            \node{\includegraphics[width=\linewidth,height=0.75\linewidth]{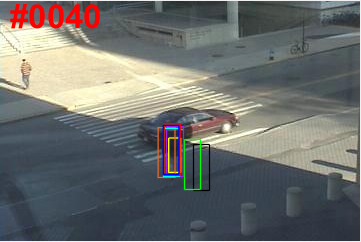}};
            \end{tikzpicture}
		\end{subfigure}
		\begin{subfigure}[normal]{0.492\linewidth}
			\begin{tikzpicture}
              \node {\includegraphics[width=\linewidth,height=0.75\linewidth]{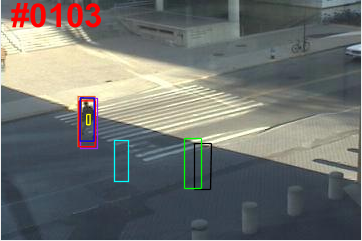}};
              \node [draw, fill=white, font=\bfseries, text=black] at (0.285\linewidth,0.295\linewidth) {\scriptsize crossing};
             \end{tikzpicture}
		\end{subfigure}
\end{subfigure}\vspace{-0.18cm}
\\
\begin{subfigure}[normal]{0.328\textwidth}
 \centering
 \begin{subfigure}[normal]{0.492\linewidth}
 			\begin{tikzpicture}
				\node{\includegraphics[width=\linewidth,height=0.75\linewidth]{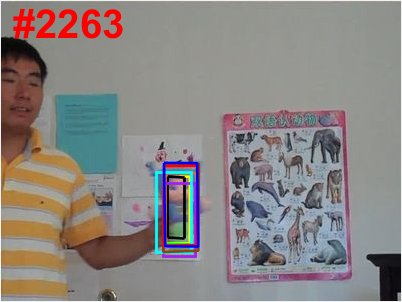}};
            \end{tikzpicture}
		\end{subfigure}
		\begin{subfigure}[normal]{0.492\linewidth}
			\begin{tikzpicture}
              \node {\includegraphics[width=\linewidth,height=0.75\linewidth]{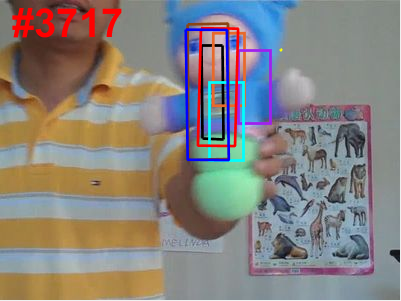}};
              \node [draw, fill=white, font=\bfseries, text=black] at (0.355\linewidth,0.295\linewidth) {\scriptsize doll};
             \end{tikzpicture}
		\end{subfigure}
\end{subfigure}
\begin{subfigure}[normal]{0.328\textwidth}
 \centering
 \begin{subfigure}[normal]{0.492\linewidth}
         \begin{tikzpicture}
			\node{\includegraphics[width=\linewidth,height=0.75\linewidth]{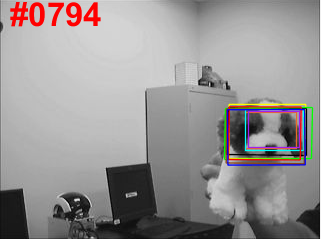}};
        \end{tikzpicture}
		\end{subfigure}
		\begin{subfigure}[normal]{0.492\linewidth}
        \centering
			\begin{tikzpicture}
              \node {\includegraphics[width=\linewidth,height=0.75\linewidth]{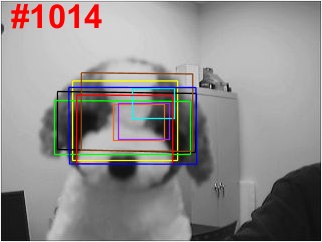}};
              \node [draw, fill=white, font=\bfseries, text=black] at (0.335\linewidth,0.285\linewidth) {\scriptsize dog1};
             \end{tikzpicture}
		\end{subfigure}
\end{subfigure}
\begin{subfigure}[normal]{0.328\textwidth}
 \centering
 \begin{subfigure}[normal]{0.492\linewidth}
         \begin{tikzpicture}
			\node{\includegraphics[width=\linewidth,height=0.75\linewidth]{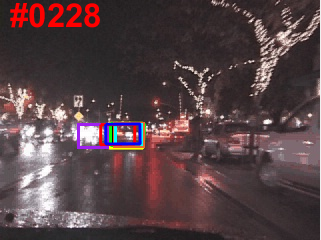}};
        \end{tikzpicture}
		\end{subfigure}
		\begin{subfigure}[normal]{0.492\linewidth}
			\begin{tikzpicture}
              \node {\includegraphics[width=\linewidth,height=0.75\linewidth]{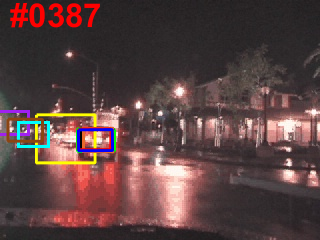}};
              \node [draw, fill=white, font=\bfseries, text=black] at (0.295\linewidth,0.290\linewidth) {\scriptsize carDark};
             \end{tikzpicture}
		\end{subfigure}
\end{subfigure}
\end{adjustbox}
\caption{Tracking results of different methods. Frame indexes are shown at the top left corner of each figure.\\
(\textcolor{black}{\textbf{---}}L1T, \textcolor{green}{\textbf{---}}MTT, \textcolor{StruckCol}{\textbf{---}}Struck, \textcolor{MILCol}{\textbf{---}}MIL, \textcolor{yellow}{\textbf{---}}IVT, \textcolor{cyan}{\textbf{---}}VTD, \textcolor{Brownish}{\textbf{---}}MTMVTLS, \textcolor{red}{\textbf{---}}MTMVTLAD, \textcolor{blue}{\textbf{---}}SMTMVT)}\vspace{-3pt}
\label{fig:cap}
\end{figure*}
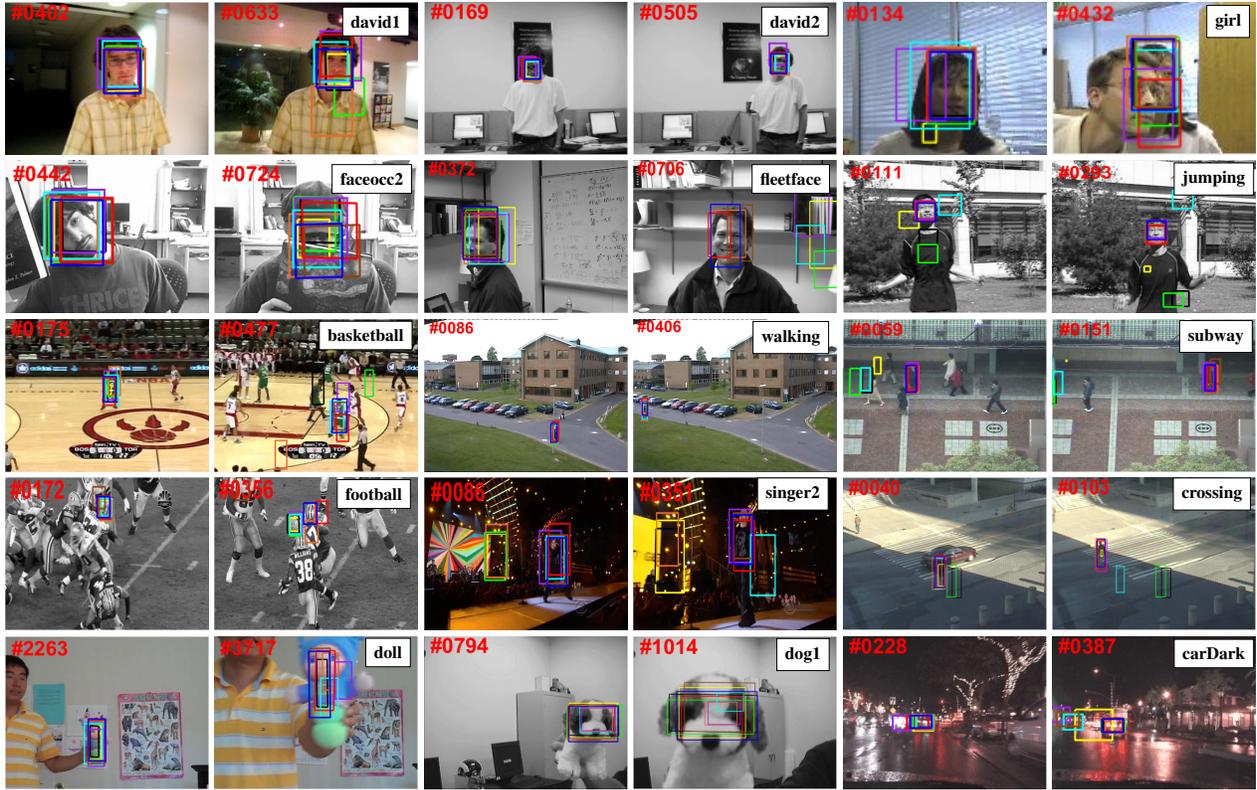

Fig.~\ref{fig:cap} demonstrates the tracking results of all compared methods on two representative frames for each of the 15 sequences.  In the \textit{david1}, \textit{david2}, \textit{girl}, \textit{faceocc2}, \textit{fleetface}, and \textit{jumping} sequences, the task is to track human faces under occlusion and scale variations. Let us take the \textit{girl} sequence as an example. IVT drifts from the target because of appearance changes. MIL and VTD are prone to drifts due to scale changes and occlusion, respectively. Struck successfully tracks the target in most frames. MTMVTLS and MTMVTLAD achieve better performance than L1T and MTT due to use of different features. SMTMVT achieves the best performance in handling the occlusion and scale variations because it retains the structure among different views.

In the \textit{basketball}, \textit{walking}, \textit{subway}, \textit{football}, \textit{singer2}, and \textit{crossing} sequences in Fig.~\ref{fig:cap}, the task is to track multiple human bodies under fast motion, rapid pose changes, and illumination variations. For instance, in the \textit{singer2} sequence, the algorithms aim to track a target with illumination variation, deformation, and rotations. IVT, L1T, MTT, and Struck quickly drift from the target mainly because of illumination changes. VTD gradually drifts from the target and loses it completely after some deformation and rotations. MIL is able to only track a part of the target without losing it. MTMVTLS and MTMVTLAD achieve good overall performance. SMTMVT achieves the best performance due to use of different views and structured representation of them.

In the \textit{doll}, \textit{dog}, and \textit{carDark} sequences in Fig.~\ref{fig:cap}, the task is to track various objects under different challenges. For instance, in the \textit{doll} sequence, the algorithms aim to track a doll with various rotations and background clutters. MTT loses the target due to the background clutter and IVT fails when the target undergoes pose changes. L1T, MIL, and Struck include much of background in the results. However, they don't lose the target since they track a part of the target throughout the frames. VTD, MTMVTLS, and MTMVTLAD achieve better performance comparing with five other methods due to incorporation of multiple features. SMTMVT produces more accurate tracking results specially when the target undergoes in-plane and out-of-plane rotations.   

For quantitative comparison, we adopt the overlap score between the tracked bounding box $r_t$ and the ground truth bounding box $r_g$ as $S=\frac{|r_t\cap r_g|}{|r_t\cup r_g|}$, where $|\cdot|$ is the number of pixels in the bounding box, $\cap$ and $\cup$ represent the intersection and union of two bounding boxes, respectively. We compute the average overlap score across all frames of each image sequence for each compared method. Table~\ref{tab:cap} summarizes the average overlap scores across all frame of each of 15 sequences for the nine compared methods. It is clear that the proposed SMTMVT method achieves the best overall performance for the tested sequences. It improves the second best method (i.e., MTMVTLAD) by $7.35\%$ in terms of the average overlap score for all 15 sequences. It ranks the best on seven sequences (e.g., \textit{david1}, \textit {girl}, \textit{subway}, \textit{singer2}, \textit{fleetface}, \textit{football}, and \textit{crossing}) and ranks the second best on four sequences (e.g., \textit{basketball}, \textit{david2}, \textit{doll}, and \textit{walking}).

\begin {table}[!htpb]
\captionsetup{justification=justified}
\caption{The average overlap score of nine compared methods for 15 sequences. Bold number in \textcolor{blue}{blue} indicates the best performance, while \textcolor{red}{red} indicates the second best.}
\centering
\scalebox{.54}{
\begin{tabular}{l||c c c c c c c c |c} 
\toprule[\heavyrulewidth]
 & \textbf{L1T} & \textbf{MTT} & \textbf{Struck} &\textbf{MIL}& \textbf{IVT} & \textbf{VTD} & \textbf{MTMVTLS} & \textbf{MTMVTLAD} & \textbf{SMTMVT}  \\ 
\midrule[\lightrulewidth]
\textbf{basketball} & 0.31  & 0.17   & 0.31   & 0.22  & 0.20  &  \textbf{\textcolor{blue}{0.71}} & 0.61  & 0.64  & \textcolor{red}{0.66} \\[0.1ex]

\textbf{david1} & 0.54  &   0.29 & 0.24 & 0.42  & 0.65 & 0.55 & \textcolor{red}{0.70} & 0.67 & \textbf{\textcolor{blue}{0.71}}\\  [0.1ex] 

\textbf{david2} & 0.79 & 0.82 & \textbf{\textcolor{blue}{0.86}} & 0.47 & 0.71 & 0.68 & 0.69 & 0.67 & \textcolor{red}{0.85} \\  [0.1ex]

\textbf{girl} & 0.70 & 0.67 & 0.70 & 0.39 & 0.21 & 0.55 & \textcolor{red}{0.72} & 0.70 & \textbf{\textcolor{blue}{0.75}} \\ [0.1ex]

\textbf{subway} & 0.16  & 0.06 & 0.63 & 0.63 & 0.16 &  0.15 & 0.62 & \textcolor{red}{0.64} & \textbf{\textcolor{blue}{0.73}} \\ [0.1ex]

\textbf{singer2} & 0.04 & 0.05 & 0.04 & 0.52 & 0.03 & 0.42 & 0.70 & \textcolor{red}{0.71} & \textbf{\textcolor{blue}{0.76}} \\ [0.1ex]

\textbf{doll} & 0.44 & 0.39 & 0.54 & 0.46 & 0.43 & 0.64 & \textbf{\textcolor{blue}{0.73}} & 0.67 & \textcolor{red}{0.71} \\  [0.1ex]

\textbf{dog1} & 0.70  & 0.68 & 0.54  & 0.53 &  \textbf{\textcolor{blue}{0.74}} & 0.59 & \textcolor{red}{0.72} & 0.69 &  0.70 \\  [0.1ex]

\textbf{faceocc2} & 0.68 & \textcolor{red}{0.74} & \textbf{\textcolor{blue}{0.78}} & 0.67 & 0.71 & 0.73 & 0.70 & 0.71 & 0.72 \\ [0.1ex] 

\textbf{fleetface} & 0.45 & 0.50 & 0.60 & 0.49 & 0.46 & 0.62 & \textcolor{red}{0.64} &  0.62  & \textbf{\textcolor{blue}{0.68}} \\  [0.1ex]

\textbf{football} & 0.55 & 0.57 & 0.53 & 0.58 & 0.55 & 0.56 & 0.35 & \textcolor{red}{0.65} & \textbf{\textcolor{blue}{0.69}} \\ [0.1ex]

\textbf{carDark} & \textcolor{red}{0.83} & 0.81 &  \textbf{\textcolor{blue}{0.89}} & 0.22 & 0.64 & 0.53 & 0.57 & 0.74 & 0.78 \\[0.1ex]

\textbf{crossing} & 0.21 & 0.25 & 0.67 & 0.71 & 0.29 & 0.31 & \textcolor{red}{0.76} & 0.73 & \textbf{\textcolor{blue}{0.78}} \\ [0.1ex]

\textbf{jumping} & 0.15 & 0.09 & 0.61 & 0.52 & 0.15 & 0.12 & \textbf{\textcolor{blue}{0.71}} & \textcolor{red}{0.70} & 0.67 \\ [0.1ex]

\textbf{walking} & \textbf{\textcolor{blue}{0.76}} & 0.64   & 0.56 & 0.54 & 0.70 & 0.60 & 0.58 & 0.60 & \textcolor{red}{0.74} \\ [0.1ex]

\midrule[\lightrulewidth]
\textbf{Average} & 0.48 & 0.45 & 0.57 & 0.49 & 0.44 & 0.51 & 0.65 & \textcolor{red}{0.68} & \textbf{\textcolor{blue}{0.73}} \\[0.1ex]
\toprule[\heavyrulewidth]
\end{tabular}
}
\label{tab:cap}
\end {table}
\vspace{-1pt}
\subsection{Experiments on CVPR2013 Tracking Benchmark}\vspace{-3pt}
We conduct the experiments on the CVPR2013 tracking benchmark \cite{wu2013online} to evaluate the performance of SMTMVT under different challenges. This benchmark consists of 50 annotated sequences. Each sequence is also labeled with attributes specifying the presence of different challenges including illumination variation (IV), scale variation (SV), occlusion (OCC), deformation (DEF), motion blur (MB), fast motion (FM), in-plane rotation (IPR), out-of-plane rotation (OPR), out-of-view (OV), background clutter (BC), and low resolution (LR). The sequences are categorized based on the attributes and 11 challenge subsets are generated. These subsets are utilized to evaluate the performance of trackers in different challenge categories. 

\begin{figure*}[t]
\centering
\captionsetup{justification=centering}
\begin{adjustbox}{minipage=\textwidth,scale=.96}
\begin{subfigure}[normal]{0.245 \textwidth}
\includegraphics[width = 0.999\linewidth]{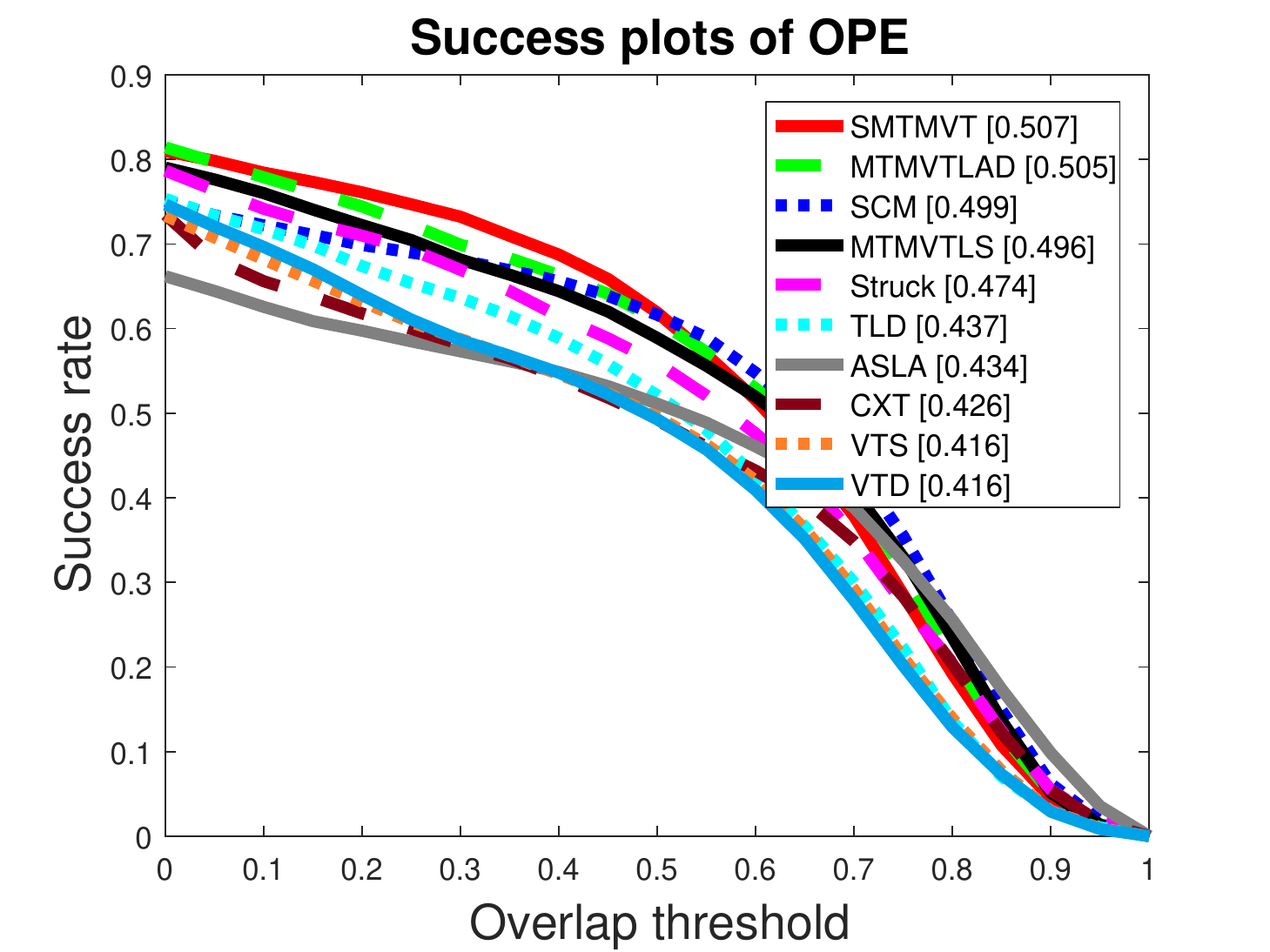}
\end{subfigure}
\begin{subfigure}[normal]{0.245 \textwidth}
\includegraphics[width =0.999\linewidth]{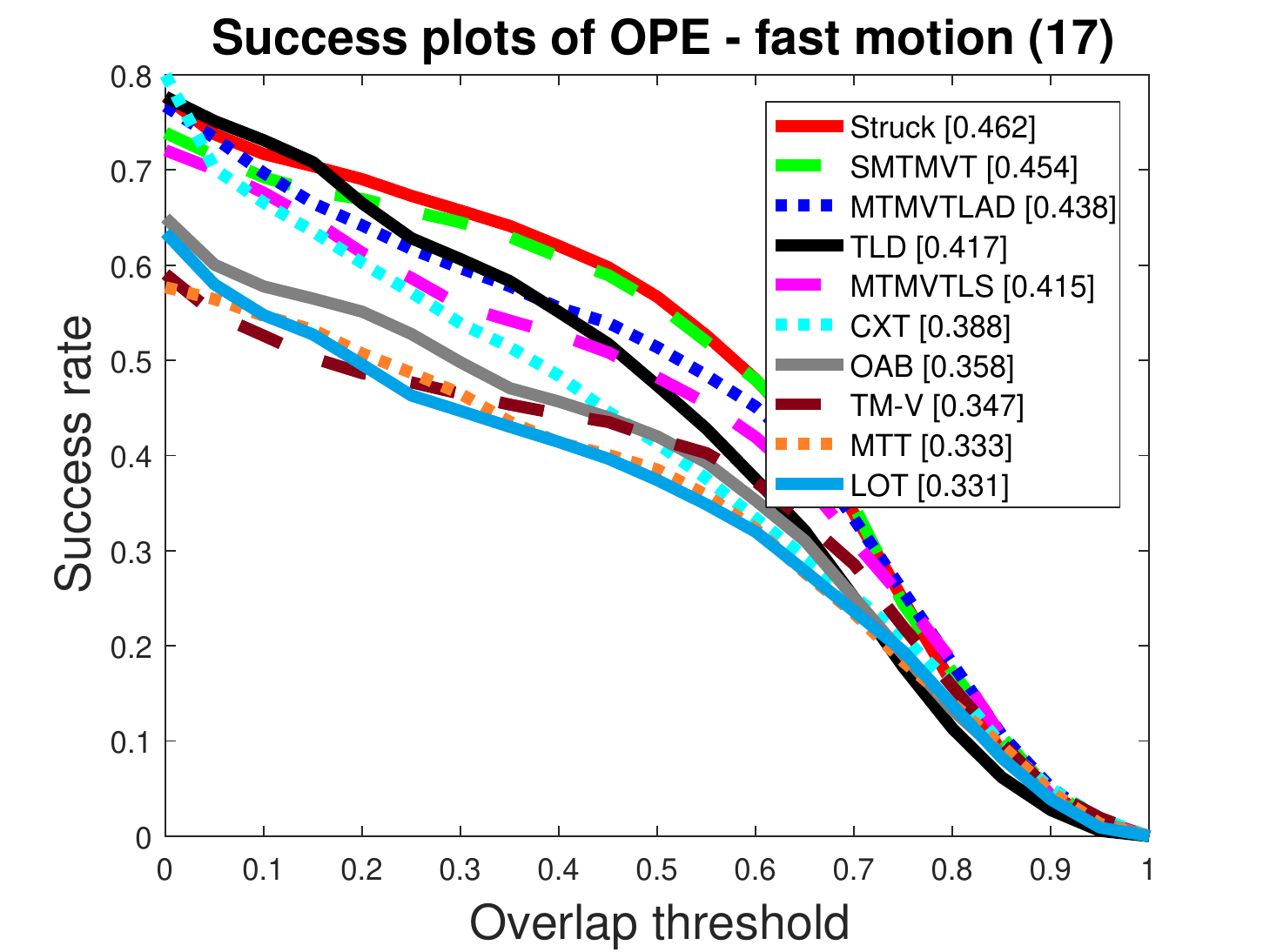}
\end{subfigure}
\begin{subfigure}[normal]{0.245 \textwidth}
\includegraphics[width =0.999\linewidth]{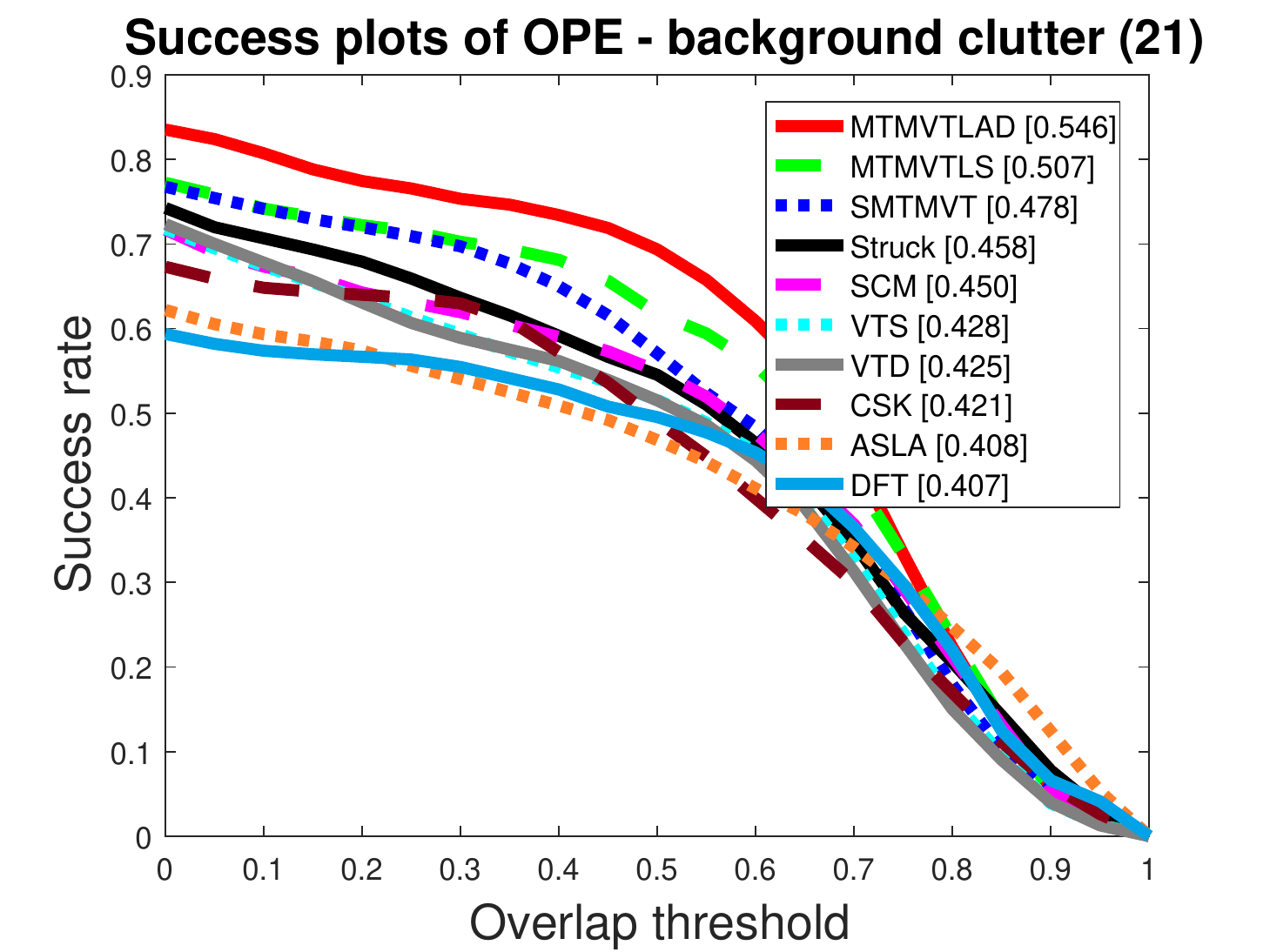}
\end{subfigure}
\begin{subfigure}[normal]{0.245 \textwidth}
\includegraphics[width = 0.999\linewidth]{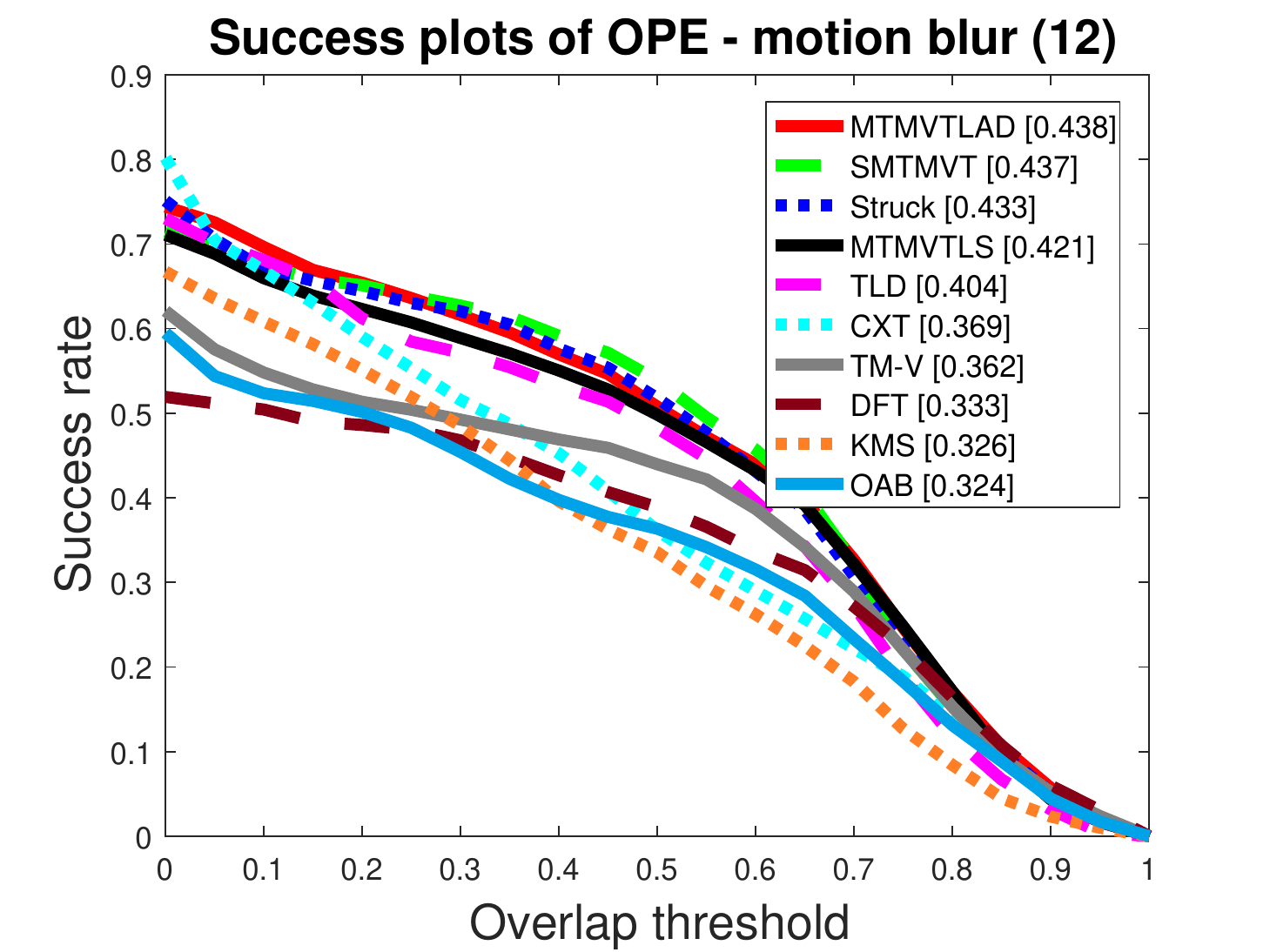}
\end{subfigure}
\\
\begin{subfigure}[normal]{0.245 \textwidth}
\includegraphics[width = 0.999\linewidth]{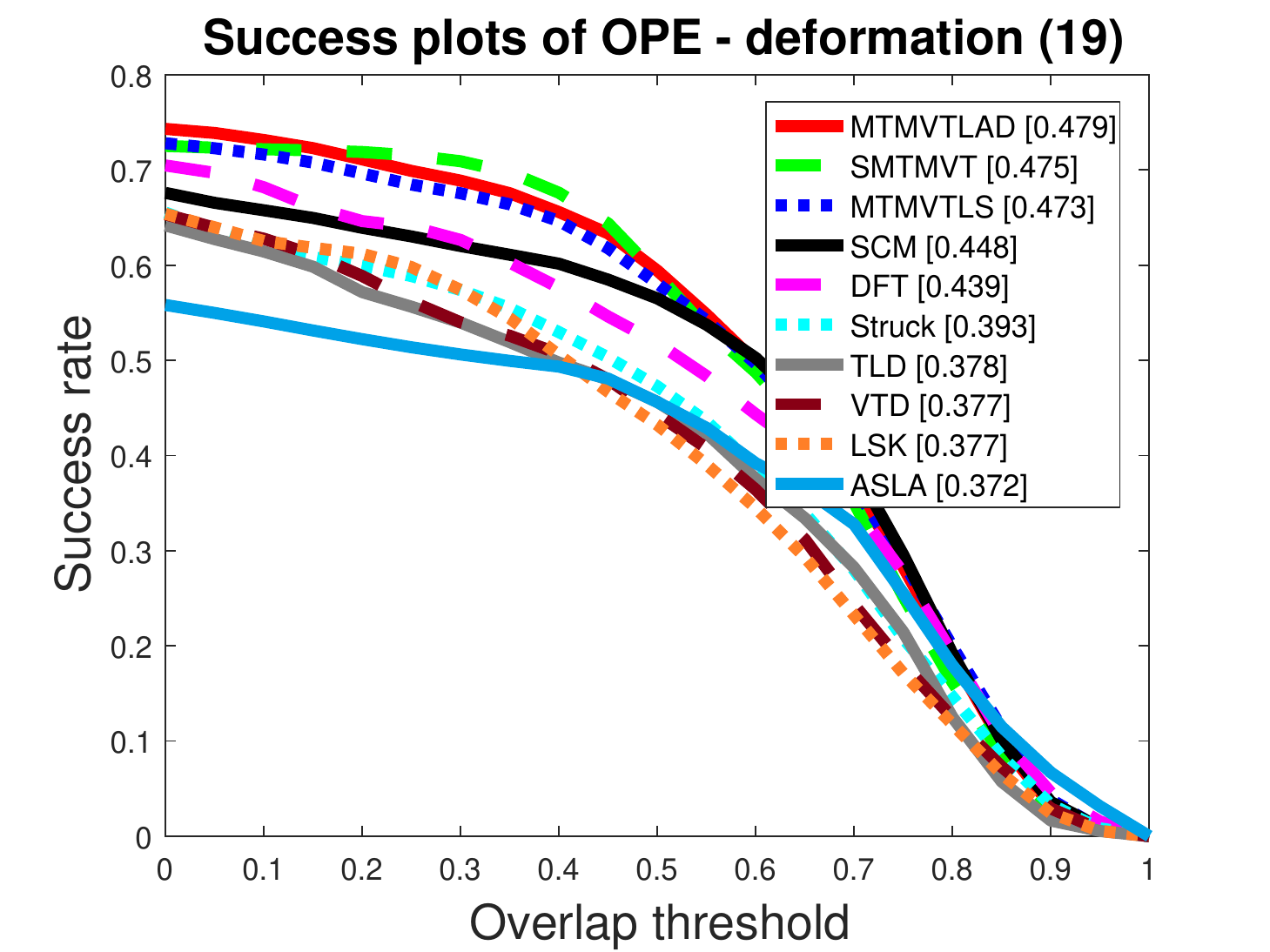}
\end{subfigure}
\begin{subfigure}[normal]{0.245 \textwidth}
\includegraphics[width =0.999\linewidth]{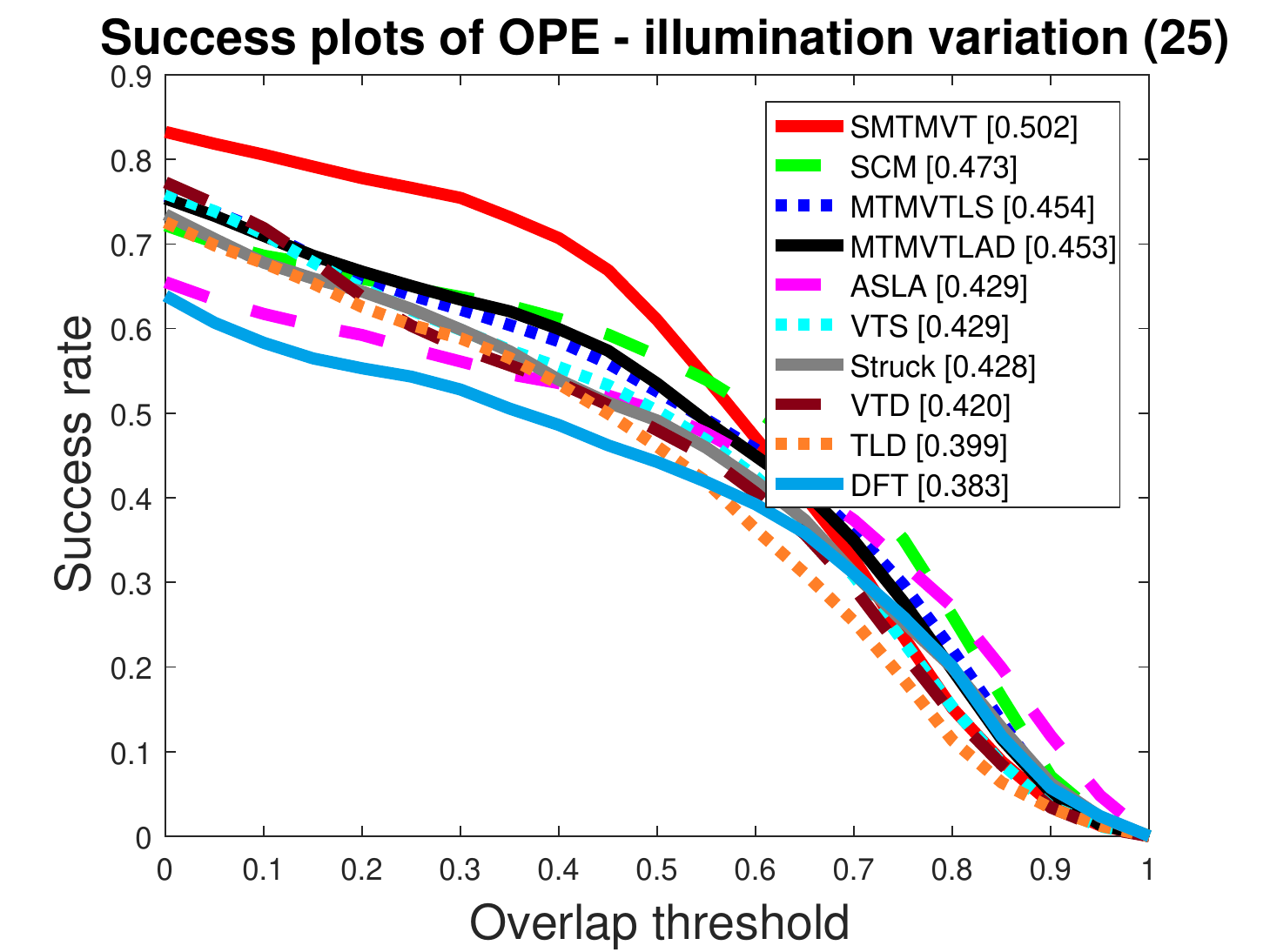}
\end{subfigure}
\begin{subfigure}[normal]{0.245 \textwidth}
\includegraphics[width =0.999\linewidth]{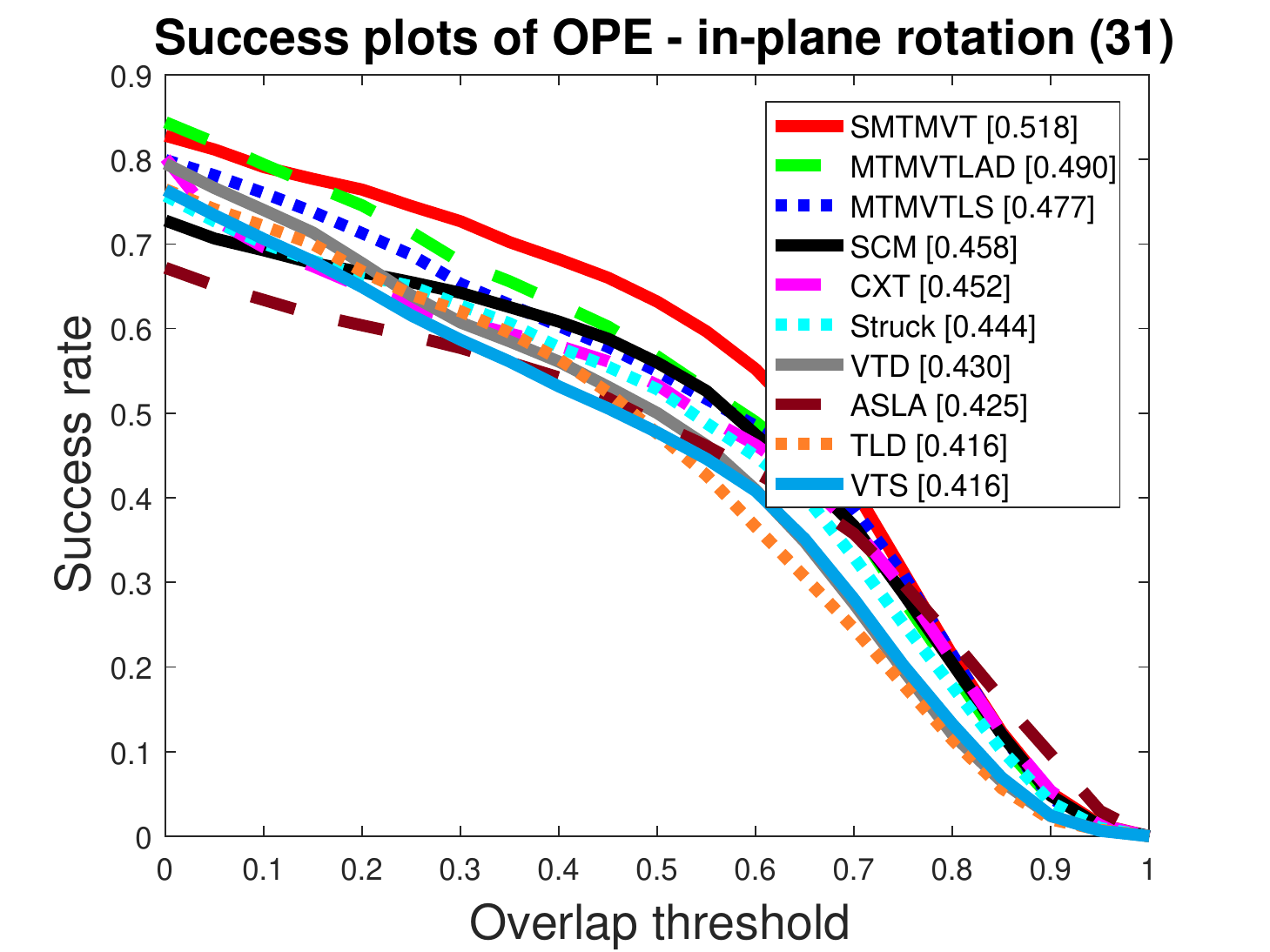}
\end{subfigure}
\begin{subfigure}[normal]{0.245 \textwidth}
\includegraphics[width = 0.999\linewidth]{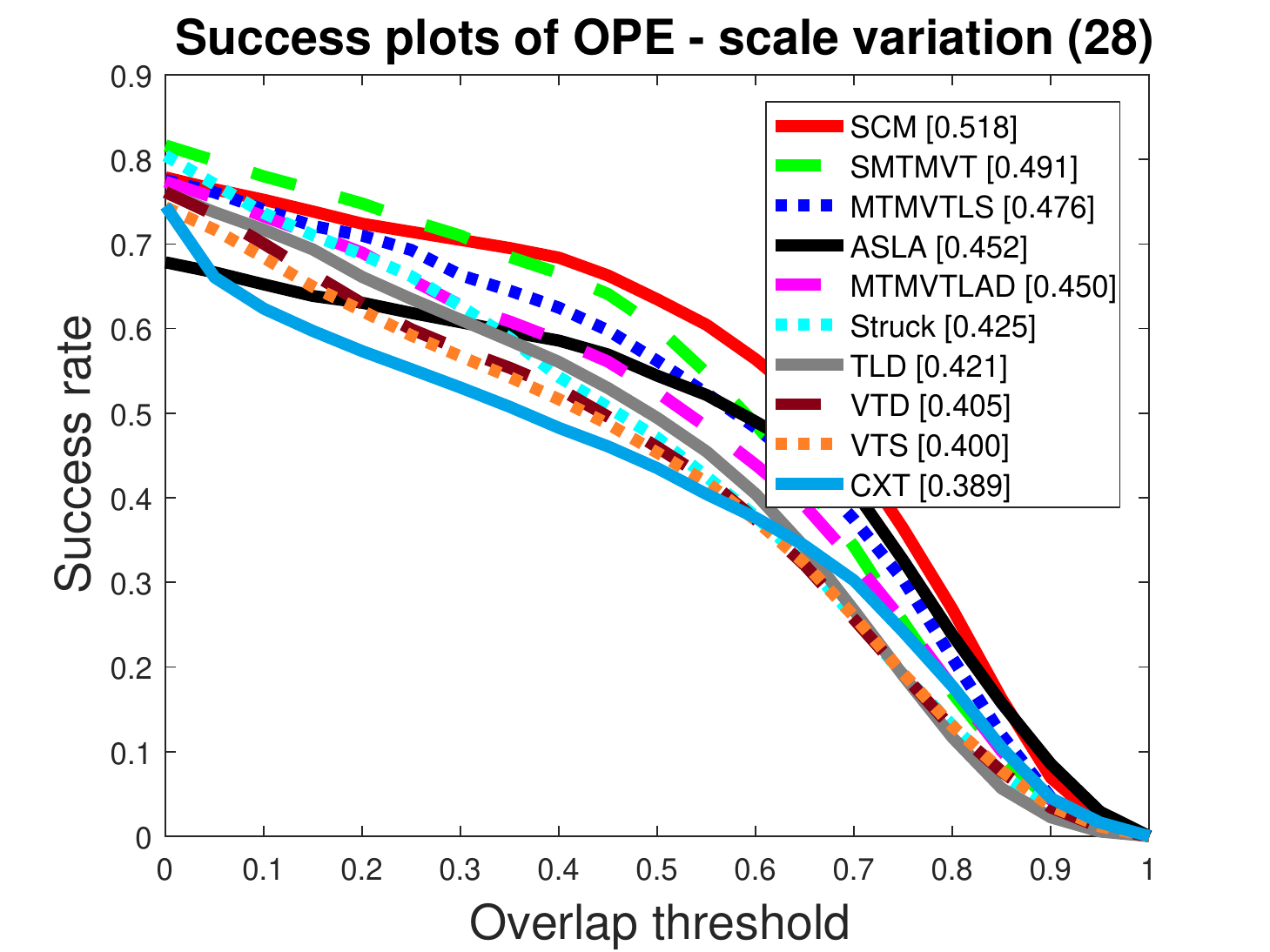}
\end{subfigure}
\\
\begin{subfigure}[normal]{0.245 \textwidth}
\includegraphics[width = 0.999\linewidth]{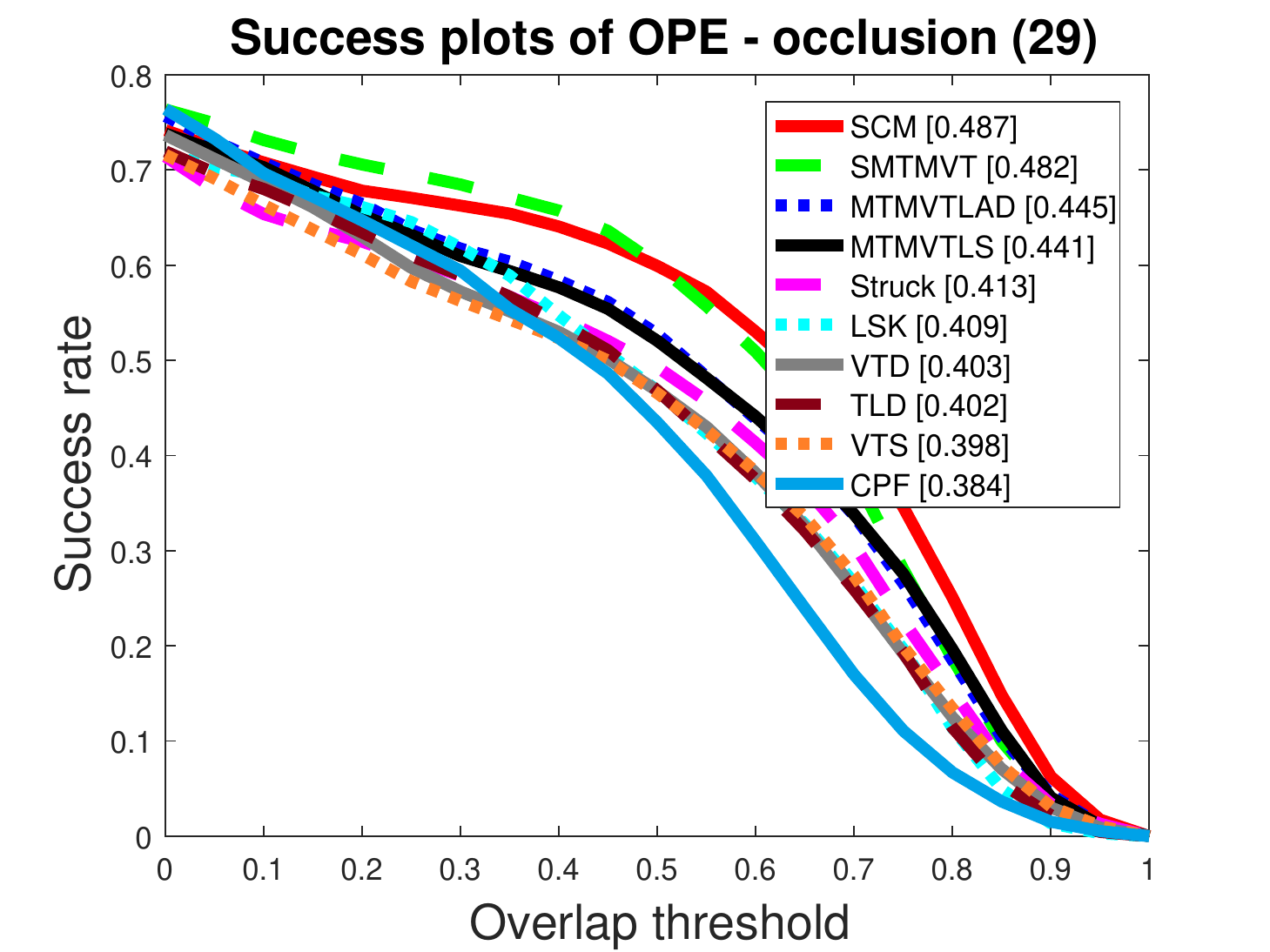}
\end{subfigure}
\begin{subfigure}[normal]{0.245 \textwidth}
\includegraphics[width =0.999\linewidth]{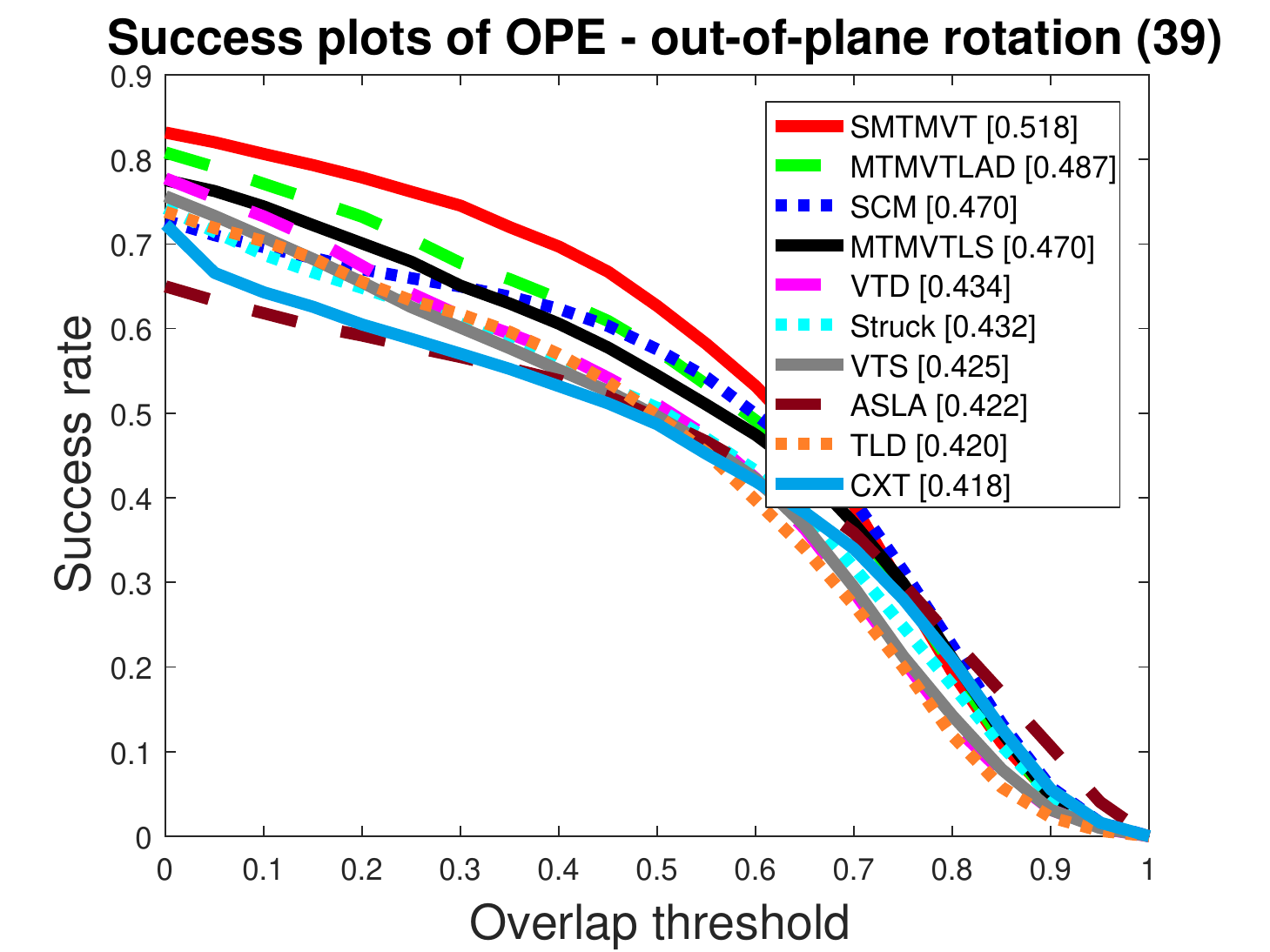}
\end{subfigure}
\begin{subfigure}[normal]{0.245 \textwidth}
\includegraphics[width =0.999\linewidth]{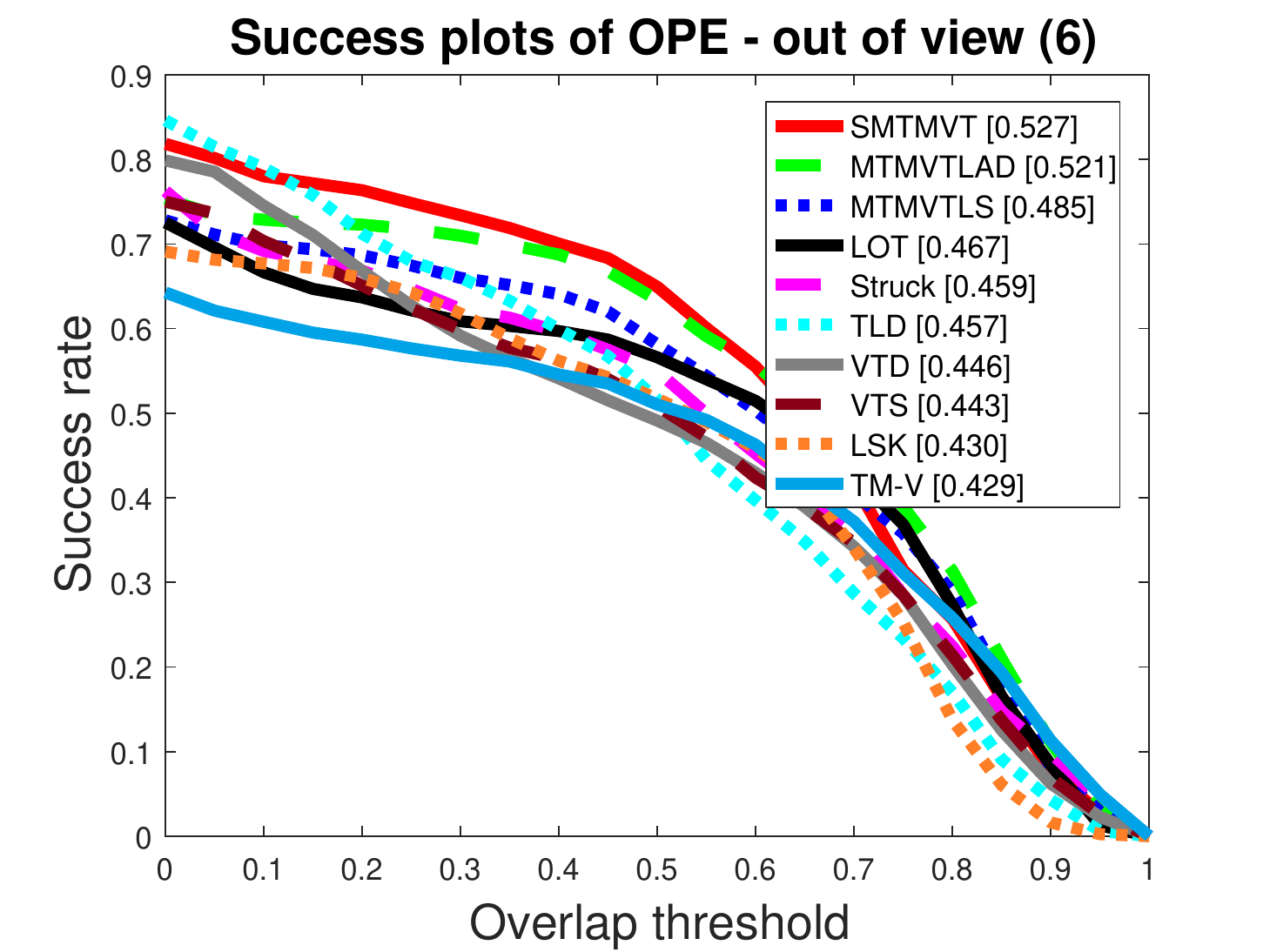}
\end{subfigure}
\begin{subfigure}[normal]{0.245 \textwidth}
\includegraphics[width = 0.999\linewidth]{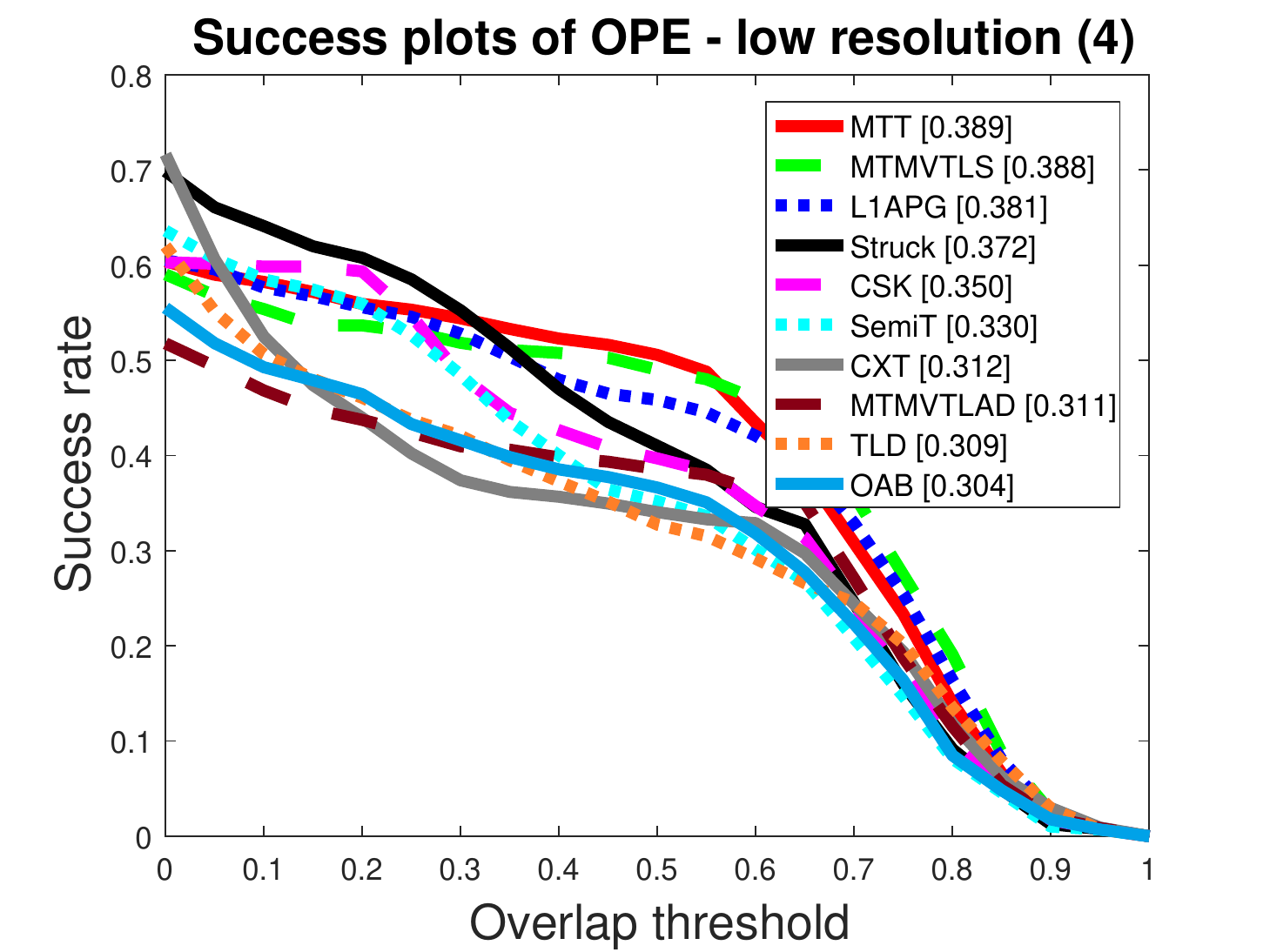}
\end{subfigure}
\end{adjustbox}
\caption{Illustration of the overall success plot and the success plot for each of 11 challenge subsets. Top 10 trackers are plotted.}\vspace{-3pt}
\label{fig:cap2}
\end{figure*}

For this benchmark dataset, there are online available tracking results for 29 trackers. In addition, we include the results of MTMVTLS and MTMVTLAD provided by the authors. Following the protocol in \cite{wu2013online}, we use the same parameters for all the sequences to produce the results for SMTMVT. We run SMTMVT to obtain the one-pass evaluation (OPE) results and compare them with the OPE results of the other 31 trackers. The OPE is conventionally used to evaluate the trackers by initializing them using the ground truth location in the first frame. We present the overall OPE success plot and the OPE success plot for each of 11 challenge subsets in Fig.~\ref{fig:cap2}. These success plots show the percentage of successful frames at the overlap thresholds ranging from 0 to 1, where the successful frames are the ones who have overlap scores larger than a given threshold. For fair comparison, we use the area under curve (AUC) of each success plot to rank the trackers.  Here, we include the top 10 of 32 trackers in each plot for clarity. The values shown in the parenthesis alongside the legends are the AUC scores. The values shown in the parenthesis alongside the titles for 11 challenge subsets are the number of video sequences in the respective subset. It is clear that SMTMVT achieves the best overall performance since it has the largest AUC score of 0.507. Also, SMTMVT ranks the best for four challenge subsets. It achieves the highest AUC score of 0.502 for IV, 0.518 for IPR, 0.518 for OPR, and 0.527 for OV. It achieves the second best for five challenge subsets (e.g., FM, MB, DEF, SV, and OCC), and the third best for BC. 
\vspace{-8pt}
\section{Conclusions}\vspace{-3pt}\label{sec:conc}
In this paper, we propose a robust SMTMVT method that uses sparse representation in the particle filter framework to track objects in challenging frame sequences. By introducing the nuclear norm regularization, we represent all views of a target candidate using the same subset of templates in the target dictionaries. We further equalize the representation coefficients of activated templates for all views. The proposed model is efficiently solved by a numerical algorithm based on the PG method. The results on 15 publicly frame sequences and the CVPR2013 tracking benchmark demonstrate that the SMTMVT method outperforms various state-of-the-art trackers.
\let\oldbibliography\thebibliography
\renewcommand{\thebibliography}[1]{%
  \oldbibliography{#1}%
  \setlength{\itemsep}{-1.7pt}%
}
\begin{spacing}{0.91}
\vspace{-3pt}
\bibliographystyle{IEEEbib}
\bibliography{camera-ready_icme2018template}

\begin{thebibliography}{10}

\bibitem{yilmaz2006object}
A.~Yilmaz, O.~Javed, and M.~Shah,
\newblock ``Object tracking: A survey,''
\newblock {\em ACM Comput. Surv.}, vol. 38, no. 4, pp. 13, 2006.

\bibitem{salti2012adaptive}
S.~Salti, A.~Cavallaro, and L.~Di~Stefano,
\newblock ``Adaptive appearance modeling for video tracking: Survey and
  evaluation,''
\newblock {\em IEEE Trans. Image Process.}, vol. 21, no. 10, pp. 4334--4348,
  2012.

\bibitem{kristan2015visual}
M.~Kristan, J.~Matas, A.~Leonardis, M.~Felsberg, L.~Cehovin, G.~Fern{\'a}ndez,
  T.~Vojir, G.~Hager, G.~Nebehay, and R.~Pflugfelder,
\newblock ``The visual object tracking vot2015 challenge results,''
\newblock in {\em ICCV}, 2015.

\bibitem{avidan2007ensemble}
S.~Avidan,
\newblock ``Ensemble tracking,''
\newblock {\em IEEE Trans. Pattern Anal. Mach. Intell.}, vol. 29, no. 2, 2007.

\bibitem{grabner2008semi}
Helmut Grabner, Christian Leistner, and Horst Bischof,
\newblock ``Semi-supervised on-line boosting for robust tracking,''
\newblock in {\em ECCV}, 2008.

\bibitem{babenko2009visual}
B.~Babenko, M-H. Yang, and S.~Belongie,
\newblock ``Visual tracking with online multiple instance learning,''
\newblock in {\em CVPR}, 2009.

\bibitem{black1998eigentracking}
M.~J. Black and A.~D. Jepson,
\newblock ``Eigentracking: Robust matching and tracking of articulated objects
  using a view-based representation,''
\newblock {\em Int. J. Comput. Vis.}, vol. 26, no. 1, pp. 63--84, 1998.

\bibitem{adam2006robust}
A.~Adam, E.~Rivlin, and I.~Shimshoni,
\newblock ``Robust fragments-based tracking using the integral histogram,''
\newblock in {\em CVPR}, 2006.

\bibitem{ross2008incremental}
D.~A. Ross, J.~Lim, R-S. Lin, and M-H. Yang,
\newblock ``Incremental learning for robust visual tracking,''
\newblock {\em Int. J. Comput. Vis.}, 2008.

\bibitem{mei2011robust}
X.~Mei and H.~Ling,
\newblock ``Robust visual tracking and vehicle classification via sparse
  representation,''
\newblock {\em IEEE Trans. Pattern Anal. Mach. Intell}, vol. 33, no. 11, pp.
  2259--2272, 2011.

\bibitem{li2011real}
H.~Li, C.~Shen, and Q.~Shi,
\newblock ``Real-time visual tracking using compressive sensing,''
\newblock in {\em CVPR}, 2011.

\bibitem{pati1993orthogonal}
Y.~C. Pati, R.~Rezaiifar, and P.~S. Krishnaprasad,
\newblock ``Orthogonal matching pursuit: Recursive function approximation with
  applications to wavelet decomposition,''
\newblock in {\em 27th Asilomar Conf. on Sig. Syst. and Compt.}, 1993.

\bibitem{shekaramiz2015block}
M.~Shekaramiz, T.~K. Moon, and J.~H. Gunther,
\newblock ``On the block-sparsity of multiple-measurement vectors,''
\newblock in {\em Sig. Process. and Sig. Process. Edu. Workshop (SP/SPE)},
  2015.

\bibitem{zhang2012robust}
T.~Zhang, B.~Ghanem, S.~Liu, and N.~Ahuja,
\newblock ``Robust visual tracking via multi-task sparse learning,''
\newblock in {\em CVPR}, 2012.

\bibitem{jia2012visual}
X.~Jia, H.~Lu, and M-H Yang,
\newblock ``Visual tracking via adaptive structural local sparse appearance
  model,''
\newblock in {\em CVPR}, 2012.

\bibitem{zhang2015structural}
T.~Zhang, S.~Liu, C.~Xu, S.~Yan, B.~Ghanem, N.~Ahuja, and M-H. Yang,
\newblock ``Structural sparse tracking,''
\newblock in {\em CVPR}, 2015.

\bibitem{yuan2012visual}
X-T. Yuan, X.~Liu, and S.~Yan,
\newblock ``Visual classification with multitask joint sparse representation,''
\newblock {\em IEEE Trans. Image Process.}, vol. 21, no. 10, pp. 4349--4360,
  2012.

\bibitem{zohrizadeh2016reliability}
F.~Zohrizadeh, M.~Kheirandishfard, K.~Ghasedidizaji, and F.~Kamangar,
\newblock ``Reliability-based local features aggregation for image
  segmentation,''
\newblock in {\em ISVC}, 2016.

\bibitem{hong2013tracking}
Z.~Hong, X.~Mei, D.~Prokhorov, and D.~Tao,
\newblock ``Tracking via robust multi-task multi-view joint sparse
  representation,''
\newblock in {\em ICCV}, 2013.

\bibitem{mei2015robust}
X.~Mei, Z.~Hong, D.~Prokhorov, and D.~Tao,
\newblock ``Robust multitask multiview tracking in videos,''
\newblock {\em IEEE Trans. Neural Netw. Learn. Syst.}, vol. 26, no. 11, pp.
  2874--2890, 2015.

\bibitem{parikh2014proximal}
N.~Parikh, S.~Boyd, et~al.,
\newblock ``Proximal algorithms,''
\newblock {\em Foundations and Trends{\textregistered} in Optimization}, vol.
  1, no. 3, pp. 127--239, 2014.

\bibitem{nesterov2005smooth}
Y.~Nesterov,
\newblock ``Smooth minimization of non-smooth functions,''
\newblock {\em Math. Prog.}, vol. 103, no. 1, pp. 127--152, 2005.

\bibitem{wu2013online}
Y.~Wu, J.~Lim, and M-H. Yang,
\newblock ``Online object tracking: A benchmark,''
\newblock in {\em CVPR}, 2013.

\bibitem{dalal2005histograms}
N.~Dalal and B.~Triggs,
\newblock ``Histograms of oriented gradients for human detection,''
\newblock in {\em CVPR}, 2005.

\bibitem{ojala2002multiresolution}
T.~Ojala, M.~Pietikainen, and T.~Maenpaa,
\newblock ``Multiresolution gray-scale and rotation invariant texture
  classification with local binary patterns,''
\newblock {\em IEEE Trans. Pattern Anal. Mach. Intell.}, vol. 24, no. 7, pp.
  971--987, 2002.

\bibitem{tan2010enhanced}
X.~Tan and B.~Triggs,
\newblock ``Enhanced local texture feature sets for face recognition under
  difficult lighting conditions,''
\newblock {\em IEEE Trans. Image Process.}, vol. 19, no. 6, pp. 1635--1650,
  2010.

\bibitem{Struck2011}
S.~Hare, A.~Saffari, and P.~H. Torr,
\newblock ``Struck: Structured output tracking with kernels,''
\newblock in {\em ICCV}, 2011.

\bibitem{babenko2011robust}
B.~Babenko, M-H. Yang, and S.~Belongie,
\newblock ``Robust object tracking with online multiple instance learning,''
\newblock {\em IEEE Trans. Pattern Anal. Mach. Intell}, vol. 33, no. 8, pp.
  1619--1632, 2011.

\bibitem{kwon2010visual}
J.~Kwon and K.~M. Lee,
\newblock ``Visual tracking decomposition,''
\newblock in {\em CVPR}, 2010.

\end{thebibliography}
\end{spacing}
\end{document}